\documentclass[lettersize,journal]{IEEEtran}
\IEEEoverridecommandlockouts
\usepackage{cite}
\usepackage{amsmath,amssymb,amsfonts}
\usepackage{algorithmic}
\usepackage{graphicx}
\usepackage{textcomp}
\usepackage{xcolor}
\usepackage{bbm}
\usepackage{bm}
\usepackage[colorlinks=true, allcolors=blue]{hyperref}
\usepackage{cite}
\usepackage{subcaption}
\usepackage{mathtools}
\usepackage{commath}
\usepackage{array}
\usepackage{xfrac}
\def\BibTeX{{\rm B\kern-.05em{\sc i\kern-.025em b}\kern-.08em
    T\kern-.1667em\lower.7ex\hbox{E}\kern-.125emX}}
\begin{document}

\newcommand{\ys}[1]{
    {\color{black} {#1}}}

\title{Aerial Robots Carrying Flexible Cables: Dynamic Shape Optimal Control via Spectral Method Model\\
\thanks{$^1$ Robotics and Mechatronics Department, Electrical Engineering,  Mathematics, and Computer Science (EEMCS) Faculty, University of Twente, 7500 AE Enschede, The Netherlands. \tt{ \footnotesize 
			\href{mailto:y.shen-2@utwente.nl}{y.shen-2@utwente.nl},
            \href{mailto:c.gabellieri@utwente.nl}{c.gabellieri@utwente.nl},
	        \href{mailto:a.franchi@utwente.nl}{a.franchi@utwente.nl}}}
\thanks{$^2$ Department of Computer, Control and Management Engineering, Sapienza University of Rome, 00185 Rome, Italy. \tt \footnotesize
			\href{mailto:antonio.franchi@uniroma1.it}{antonio.franchi@uniroma1.it}
}
\thanks{This work was partially funded by the European Commission Horizon Europe Framework under project Autoassess (101120732) and MSCA PF Flyflic (101059875), and by the China Scholarship Council (202206290005).}
}

\author{\IEEEauthorblockN{Yaolei Shen$^{1}$~\IEEEmembership{Student Member~IEEE}},
\and
\IEEEauthorblockN{Antonio Franchi$^{1,2}$~\IEEEmembership{Fellow~IEEE}},
\and
\IEEEauthorblockN{Chiara Gabellieri$^{1}$~\IEEEmembership{Member~IEEE}}

}

\newcommand{\revision}[1]{\textcolor{black}{#1}}

\maketitle

\begin{abstract}
In this work, we present a model-based optimal boundary control design for an aerial robotic system composed of a quadrotor carrying a flexible cable. The whole system is modeled by partial differential equations (PDEs) combined with boundary conditions described by ordinary differential equations (ODEs). The proper orthogonal decomposition (POD) method is adopted to project the original infinite-dimensional system on a \textcolor{black}{finite low-dimensional space} spanned by orthogonal basis functions. Based on such a reduced order model, nonlinear model predictive control (NMPC) is implemented online to realize \revision{both position and} shape trajectory tracking of the flexible cable in an optimal predictive fashion. The proposed \textcolor{black}{POD-based} reduced modeling and optimal control paradigms are verified in \textcolor{black}{simulation using} an accurate high-dimensional FDM-based model and \textcolor{black}{experimentally using a real quadrotor and a  cable}. \textcolor{black}{The results show the the viability of the POD-based predictive control approach (allowing to close the control loop on the full system state) and its superior performance compared to a optimally tuned PID controller (allowing to close the control loop on the quadrotor state only).} 
\end{abstract}


\section{Introduction}
\IEEEPARstart{R}{ecently,} uncrewed aerial vehicles (UAVs), also known as aerial robots,  have attracted significant attention due to their huge potential applications in areas such as search and rescue, agriculture, transportation, or even inspection and \revision{maintenance}. Among this large number of scenarios, aerial vehicles may need to interact with deformable objects, for example, when transporting suspended cargo through a cable \cite{maza2010multi,villa2020survey,brandao2022side,bernard2009generic,tagliabue2017collaborative} or when performing refueling operations using flexible pipes \cite{ro2010modeling,liu2017modeling,liu2022ann,song2022dynamics}. \ys{Additionally, some industrial applications such as tethered powered aerial robotics \cite{Jain2022tethered} and fire-fighting with drones \cite{Aerones2018} 
show similar configurations, where the aerial robots are tethered with flexible cables or hoses}. In these cases, an aerial vehicle manipulates a flexible cable through just one contact point. Due to the flexible characteristic of the cable, modeling and control for such a soft aerial manipulation system are different from those used in conventional rigid aerial manipulation \cite{ollero2021past,tagliabue2019robust}.


For analyzing and synthesizing these aerial vehicle-cable systems, several mathematical models have been proposed in the robotics community, and  different control laws and planning methods were designed based on such models. \revision{Models of an aerial vehicle-cable system found in the literature} can be generally divided into five categories: single rigid pendulum model, single elastic pendulum model, series of rigid links model, series of elastic links model, and distributed parameter model.

\noindent \emph{1) Single rigid pendulum}: Palunko et al. consider a system composed of a quadrotor and a cable-suspended load, they model it as a single pendulum attached to the aerial vehicle, in which the cable is mass-less and non-elastic and the load is represented as a point mass \cite{palunko2012trajectory}. Based on this 3D single pendulum, they adopt a dynamic programming approach to generate a swing-free trajectory. The same model is also used in \cite{pizetta2015modelling}, where the motion of the pendulum is regarded as a disturbance, and feedback linearization control is used. Furthermore, this single rigid pendulum model is used to solve planning and control problems in single multi-rotor suspended payload transportation \cite{son2018model,son2020real} and quadrotor cooperative transportation \cite{gassner2017dynamic,li2021cooperative,li2023nonlinear}. In \cite{sreenath2013geometric} and \cite{tang2015mixed}, the system is modeled using two subsystems, one in which the cable is taut and in which it is slack, each subsystem is proved to be differentially flat, which is used for trajectory generation. Hence, a hierarchical controller is designed to track the planned trajectories.  

\noindent \emph{2) Single elastic pendulum}: Cable models that neglect the bending but allow for changes in the length via elasticity have been also adopted in the literature. In \cite{kotaru2017dynamics} and \cite{foehn2017fast},  a point mass is suspended below a single quadrotor, while a rigid body manipulated by multiple quadrotors through elastic cables is considered in~\cite{goodman2022geometric} and~\cite{goodman2023geometric}. Controllers for tracking the payload's pose are designed on a simplified model disregarding elasticity and are proven to work in the elastic case as well.  
In~\cite{tognon2018aerial, gabellieri2020study, gabellieri2023equilibria} admittance-based control not relying on direct communication is used to regulate the pose of a payload suspended below elastic cables.

\noindent \emph{3) Series of rigid links}: To include the flexibility of the cable in the model, Goodarzi et al. use a finite series of rigid links (multiple pendulum model) to describe the behavior of the cable, inextensible mass-less links are connected by passive spherical joints, where the mass is concentrated \cite{goodarzi2015geometric}. A similar discretized model is also used to model the motion of the aerial refueling hose \cite{ro2010modeling}. In \cite{kotaru2018differential,kotaru2020multiple}, Kotaru et al. prove the differential flatness of different systems composed by aerial vehicles manipulating a cable modeled with a series of rigid links, and they use LQR to control these systems. Instead of using lumped mass link models, Quisenberry et al. use distributed mass links to model the cable \cite{quisenberry2006discrete}.

\noindent \emph{4) Series of elastic links}: Series of elastic links are adopted to consider the flexibility, elasticity, and length variation of the cable in the tethered aerial system \cite{song2022dynamics}.  Based on this lumped-mass model, \cite{song2022dynamics} uses dynamic surface control (DSC) to control the position of a suspended load. A similar cable model has been considered in \cite{gabellieri2023differential} for a two-robot cable system.

\noindent \emph{5) Distributed parameters model}: Different from the discretized models seen so far, Liu et al. use a distributed parameter model represented by partial differential equations to describe the motion of the hose in an aerial refueling system in the longitudinal plane \cite{liu2017modeling}, and a backstepping approach is adopted to suppress the vibration of the hose. In \cite{liu2022ann}, Liu et al. use an adaptive neural network to handle the model uncertainties.


To summarize, while single pendulum models may simplify well the dynamics of the aerial cable-suspended manipulation system when a payload of considerable mass forces the cable into a straight configuration, more complex models have been used in the literature to capture the flexibility of the cable when the task requires it. Discrete models fail to capture accurately the dynamics of the continuously flexible cable, unless a very high number of discrete elements is used, making them computationally inefficient. Distributed-parameter models have been proposed in the literature for vibration suppression during payload transportation, but not for highly dynamic tasks where the cable shape is controlled.

\begin{figure}
\centering
\subfloat[Robotic platform]{%
  \includegraphics[clip,height=3.8cm]{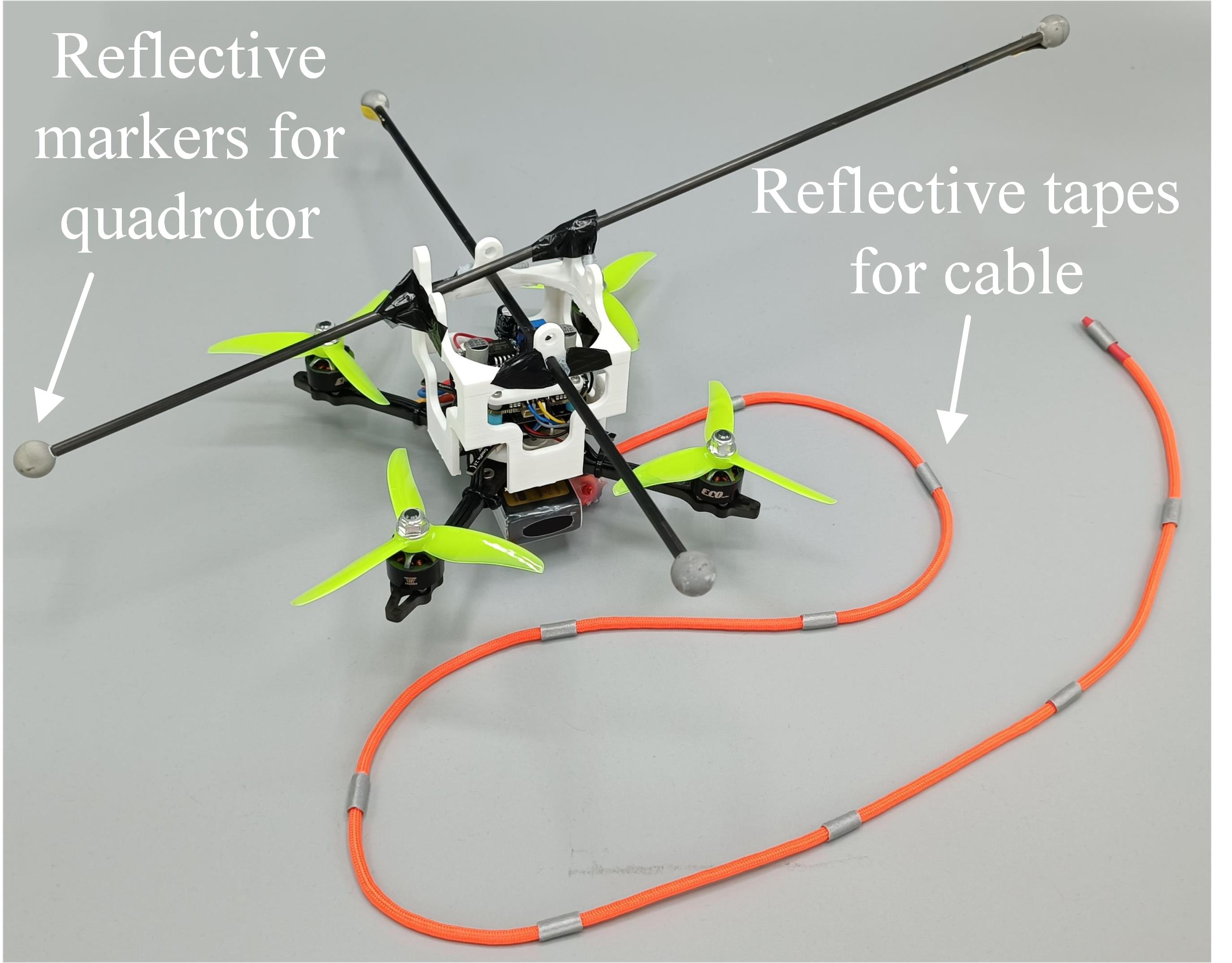}%
}
\subfloat[System configuration]{%
  \includegraphics[clip,height=4.3cm]{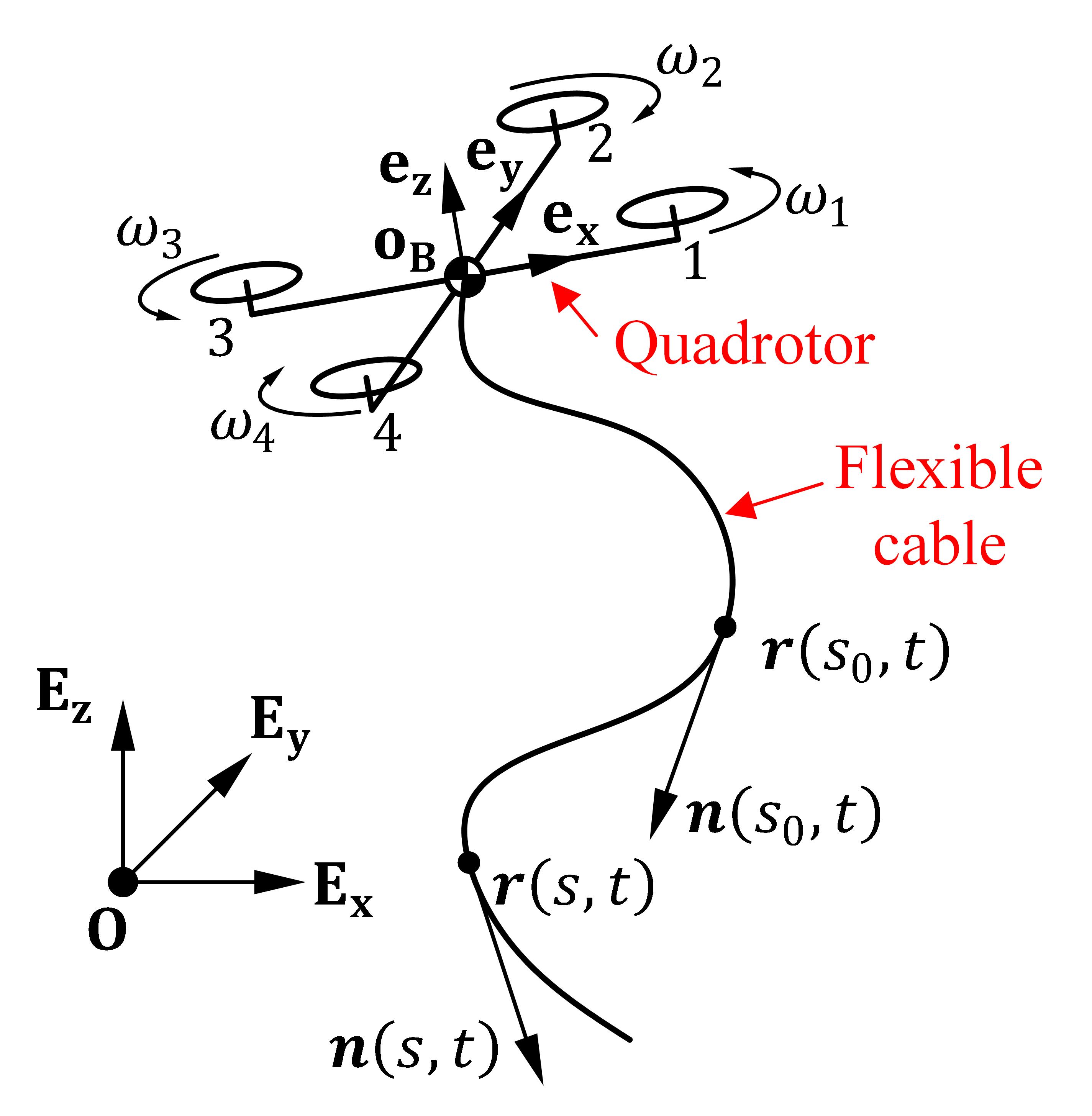}%
}
\caption{(a): 10\,cm quadrotor equipped attached to a cable, which is measured through reflective tape; (b): schematics of the quadrotor-cable system and its configuaration.}
\label{fig:1}
\end{figure}

In this work, we focus on a system composed of a quadrotor carrying a flexible and extensible cable, which is illustrated in Figure~\ref{fig:1}. This work aims to control the motion of the quadrotor to let the passive cable \revision{follow a given path while generating} different kinds of shapes. A distributed parameter model, formulated by PDEs, is used to describe the motion of the cable in 3D space. The 6-DoF motion of the quadrotor is described by ODEs, which represent one of the two boundary conditions of the cable model. For controlling this kind of PDE-ODEs hybrid system, the \emph{Proper Orthogonal Decomposition} (POD) method is adopted to reduce the order of the original system. That is done by projecting the original infinite-dimensional system on a \revision{finite low-dimension space} spanned by orthogonal basis functions. After that, such reduced-order model is adopted as the reference system for a nonlinear model predictive control to track the cable’s shape trajectory. To our best knowledge, the contributions of this work are the following:
\begin{itemize}
\item a novel distributed-parameter mathematical model is presented to describe and simulate the evolution of the quadrotor-cable system accurately;
\item the spectral-decomposition method is used in this work for the first time for deriving a simplified model of the aerial vehicle-cable system that is used for the control design;
\item this work considers and solves for the first time the cable’s shape-tracking control problem in aerial vehicle-cable systems, and \revision{the proposed method is validated through simulations against a finite high-dimensional FDM-based model and with real experiments.}
\end{itemize}

This paper is organized as follows: Section~\ref{sec:model} presents the derivation of the system dynamics. In Section~\ref{sec:model_reduced}, the model order reduction process is derived based on the spectral method. In Section~\ref{sec:control},  NMPC-based trajectory tracking is presented. Section~\ref{sec:sims} shows numerical simulations that validate the proposed controller\textcolor{black}{, while experimental results are in Section~\ref{sec:real}.} Concluding remarks are listed in section~\ref{sec:conclusions}.

\section{Mathematical Model}\label{sec:model}

\subsection{System Description}

The quadrotor-cable system in this work is illustrated in Figure~\ref{fig:1}: one endpoint of the flexible cable is attached to the center of mass (CoM) of the quadrotor. During the flight, the motion and deformation of the flexible cable are affected by its inertia, gravity, external aerodynamic forces, and the external force exerted by the robot. The work considers the following simplifying assumptions:
\begin{itemize}
\item The air medium is homogeneous, which means the density of the air is the same everywhere;
\item The air medium is static, and the effect of the airflow generated by the quadrotor’s propellers on the cable is neglected;
\item The cable is perfectly flexible, namely, there is no torque generated when the cable is bending or twisting.
\end{itemize}

\subsection{System Modeling}

\emph{1) Kinematics}: In this section, the kinematics of the quadrotor-cable system is described. Let \{$\mathbf{E_x}={\rm [1,0,0]^T}$,\,$\mathbf{E_y}={\rm [0,1,0]^T}$,\,$\mathbf{E_z}={\rm [0,0,1]^T}$\} be a fixed orthonormal basis for the three-dimensional Euclidean space.  The cable is defined as a set of elements called material points and its configuration as a curve in the world frame (absolute inertial frame), which is indicated as $\mathbf{OE_x E_y E_z}$. 

In this work, the subscripts $(\cdot)_\star$ and $(\cdot)_{\star\star}$ represent the operations $\partial (\cdot)/\partial \star$ and $\partial^2 (\cdot)/\partial \star^2$, respectively.  \ys{The first and second order total derivative of time $d(*)/dt$ are $d(*)^2/d^2t$ are represented as $\Dot{(*)}$ and $\Ddot{(*)}$, respectively.} \ys{The curve describing the cable geometry is defined as a map $\bm{r}: (s,t) \in [0,L]\times[0,+\infty] \mapsto \bm{r}(s,t) \in \mathbbm{{R}}^3$ w.r.t the world frame, where $s$ is the curvilinear coordinate of the curve, $t$ represents time, and $L$ equals the arc length of the curve when the cable is at rest (there are no strains along the cable). The two endpoints of the cable $\bm{r}(0,\cdot)$ and $\bm{r}(L,\cdot)$ are called head tip and tail tip, respectively.} The velocity and acceleration of the material point are denoted as $\bm{r}_t(s,t)$ and $\bm{r}_{tt}(s,t)$, respectively. 
Because\ys{the head tip} of the cable coincides with the CoM of the quadrotor, the position of the quadrotor's CoM\ys{$\bm{p}_B$} with respect to the world frame equals $\bm{r}(0,t)$. The attitude of the quadrotor is expressed by the rotation matrix $\mathbf{R}_B\in {\rm SO(3)}$ transforming coordinates from  $\mathbf{OE_x E_y E_z}$ to the quadrotor's body-fixed frame $\mathbf{o_B e_x e_y e_z}$ which is shown in Figure~\ref{fig:1}. In this work, we use Tait–Bryan angles \{$\theta _z$,$\theta _y$,$\theta _x$\} following the rotation order ${\rm Z \rightarrow Y \rightarrow X}$ to represent $\mathbf{R}_B$. The angular velocity of the $\mathbf{o_B e_x e_y e_z}$ w.r.t. $\mathbf{OE_x E_y E_z}$ expressed in $\mathbf{o_B e_x e_y e_z}$ is denoted with $\bm{\omega}_B\in \mathbbm{R}^3$. Its relationship with the rate of change of the Tait–Bryan angles is:
\begin{equation}
    \bm{\omega}_B=
        \mathbf{T}^{-1}
    [
         \Dot{\theta}_x,
         \Dot{\theta}_y,
         \Dot{\theta}_z]^{{\rm T}},
\label{eq:1}
\end{equation}
where $\bm{{\rm T}}$ is the transformation matrix from the body rate in the body-fixed frame to the rate of Tait–Bryan angles.

\emph{2) Constitutive law}: The contact force $\bm{n}(s,t)\in \mathbbm{R}^3$ that the material point in position $\bm{r}(s,t)$ undergoes to due to the action of the cable portion $[s,\,L]$ can be related through material constitutive laws to the mechanical strains. For more detailed information about the nonlinear elasticity of extensible cable please refer to \cite{Antman:1250280}. In this work, the cable is considered a perfectly flexible cable, so elongation, expressed as in \eqref{eq:2}, is the only type of strain involved in the model: 
\begin{equation}
    \varepsilon(s,t)=\frac{\varepsilon(s,t)ds}{ds}=\frac{\left\lVert  \bm{r}_s(s,t) ds \right\rVert _2  -ds}{ds}=\left\lVert \bm{r}_s(s,t) \right\rVert _2-1,
\label{eq:2}
\end{equation}%
\revision{where $ds$ is a differential element of the cable length, when this cable element experiments a strain $\varepsilon(s,t)$, $ds$ elongates to the actual cable differential element length $\left\lVert \bm{r}_s(s,t) ds \right\rVert _2$.} Due to perfect flexibility, the internal force $\bm{n}(s,t)$ is tangent to the curve at $\bm{r}(s,t)$ and is expressed as 

\begin{equation}
    \bm{n}(s,t)=n(s,t)\frac{\bm{r}_s(s,t)}{\left\lVert \bm{r}_s(s,t) \right\rVert _2}
\label{eq:3}
\end{equation}
\noindent where $n(s,t)$ is the cable’s tension at point $(s,t)$. In this work, we consider the cable is linearly elastic, and so the tension $n(s,t)$ is denoted as
\begin{equation}
    n(s,t)=EA\cdot\varepsilon(s,t),
\label{eq:4}
\end{equation}%
where $E$ and $A$ are Young's modulus and the cross-sectional area of the cable, respectively.

\emph{3) Aerodynamic load and gravitational force}: During motion, the cable is subject to two different external forces: aerodynamic drag and gravitational force. The aerodynamic drag $\bm{d}(s,t)$ acting on the material point with configuration $s$ at time $t$ (which we indicate as point $(s,t)$) is proportional to the square of the velocity vector and is directed in the opposite direction. The expression of the aerodynamic drag is given by

\begin{equation}
    \bm{d}(s,t)=-\rho _a c_d \bm{r}_t^{\rm T} \bm{r}_t \frac{\bm{r}_t}{\left\lVert \bm{r}_t \right\rVert _2},
\label{eq:5}
\end{equation}
in which $\rho _a$ is the air density,  $c_d$ is the aerodynamic drag coefficient, and we omitted the dependency of $\bm{r}$ on $s$ and $t$. The gravitational force $\bm{g}(s,t)$ that the cable undergoes at point $(s,t)$ is:
\begin{equation}
    \bm{g}(s,t)=-\rho _c Ag \mathbf{E_z},
\label{eq:6}
\end{equation}
in which $\rho _c$ is the density of the
cable and $g={\rm 9.8 \, m\cdot s^{-2}}$ is the value of acceleration of gravity.

\emph{4) Equations of motion}: Consider now a segment $[s_0,s]$ of the cable. According to the law of conservation of momentum, the motion of the selected segment follows the following  integral form: 
\begin{equation}
\begin{aligned}
    \bm{n}(s,t)-\bm{n}(s_0,t)+\int_{s_0}^{s} [\bm{d}(\xi,t)+\bm{g}(\xi,t)] \,d\xi \\
    = \frac{d}{dt} \int_{s_0}^{s} \rho _c A \bm{r}_t(\xi,t) \,d\xi
\end{aligned}
\label{eq:7}
\end{equation}

After differentiating~\eqref{eq:7} with respect to $s$, the equations of motion of the cable are obtained as 
\begin{equation}
    \bm{n}_s(s,t)+\bm{d}(s,t)+\bm{g}(s,t)=\rho_c A \bm{r}_{tt}(s,t)
\label{eq:8}.
\end{equation}
The equations of motion of the quadrotor are derived using the Newton-Euler approach:
\begin{equation}
    \left[
        \begin{array}{c}
            m_B \bm{r}_{tt}(0,t) \\
            \bm{{\rm J}}_B \Dot{\bm{\omega}}_B \\
        \end{array}
    \right]
    =-
    \left[
        \begin{array}{c}
            m_B g \mathbf{E_z} \\
            \bm{{\rm J}}_B \times {\rm \bm{{\rm J}}}_B \bm{\omega}_B \\
        \end{array}
    \right]
    +
    \left[
        \begin{array}{c}
            \bm{f}^W \\
            \bm{\tau}^B \\
        \end{array}
    \right],
\label{eq:9}
\end{equation}
 in which $m_B$ is the mass of the quadrotor, $\bm{{\rm J}}_B$ is ${\rm 3 \times 3}$ rotational inertia matrix w.r.t. $\bm{{\rm o_B}}$ and expressed in the frame $\bm{{\rm o_B e_x e_y e_z}}$,  $\bm{f}^W \in \mathbbm{R}^3$ is the sum of the input and external forces applied to the quadrotor’s CoM, expressed in the world frame, and $ \bm{\tau}^B$ the total input torque, expressed in the body-fixed frame. The expression of $\bm{f}^W$ is shown in~\eqref{eq:10} and is composed of two parts: the thrust input force generated by the four propellers and the interaction force generated by the flexible cable: 
\begin{equation}
    \bm{f}^W=\mathbf{R}_B \sum\nolimits_{i=1}^{4} \bm{f}_T^i + \bm{n}(0,t)
\label{eq:10}
\end{equation}
 where $\bm{f}_T^i=c_T^i \omega_i^2 \bm{{\rm e_z}}$ is the thrust force generated by the ${\rm i^{th}}$ propeller\ys{in the body-fixed frame}, with $c_T^i$ and $\omega_i$ the thrust coefficient and intensity of rotational velocity of the ${\rm i^{th}}$ propeller, respectively.\ys{$\bm{n}(0,t)$ is the internal force (tension) that the cable head tip undergoes to, which is also the interaction force that the cable applies on the quadrotor.} 

The expression of $\bm{\tau}^B$ is shown as~\eqref{eq:11}, where $\bm{r}_p^i \in \mathbbm{R}^3$ is the position of the ${\rm i^{th}}$ propeller in the quadrotor's body-fixed frame and $c_{\tau}^i$ is the drag coefficient \cite{hamandi2021design}.
\begin{equation}
    \bm{\tau}^B=\sum\nolimits_{i=1}^{4} [\bm{r}_p^i \times \bm{f}_T^i + (-1)^i c_{\tau}^i \omega_i^2 \bm{{\rm e_z}}].
\label{eq:11}
\end{equation}

After substituting \eqref{eq:2}-\eqref{eq:6} into \eqref{eq:8}, the dynamics of the quadrotor-cable system can be described by  
\begin{equation}
    \bm{n}_s(s,t)-\rho_a c_d \bm{r}_t \cdot \left\lVert \bm{r}_t \right\rVert _2 -\rho_c A 
    g\bm{{\rm E_z}}
    =\rho_c A \bm{r}_{tt}
\label{eq:12}
\end{equation}
for ${\forall s \in (0,L], t \in [0,\infty)}$, combined with the boundary conditions~\eqref{eq:13}-\eqref{eq:15}.\ys{The boundary condition~\eqref{eq:15} means the tail tip of the cable is not strained.}
\begin{align}
    &m_B \bm{r}_{tt}(0,t) =-m_B g \bm{{\rm E_z}}+\mathbf{R}_B \sum\nolimits_{i=1}^{4} \bm{f}_T^i + \bm{n}(0,t)
\label{eq:13}\\
    &\bm{{\rm J}}_B \Dot{\bm{\omega}}_B =\sum_{i=1}^{4} [\bm{r}_p^i \times \bm{f}_T^i + (-1)^i c_{\tau}^i \omega_i^2 \bm{{\rm e_z}}]-\bm{\omega}_B \times {\rm \bm{{\rm J}}}_B \bm{\omega}_B
\label{eq:14}\\
    &||\bm{r}_s(L,t)||_2 =1.
\label{eq:15}
\end{align}

\subsection{Finite Difference Method}\label{subs:fdm}

The finite difference method (FDM) is adopted to simulate  the quadrotor-cable system by solving numerically the partial differential equations (PDEs)~\eqref{eq:12} combined with boundary conditions in~\eqref{eq:13}-\eqref{eq:15} and initial conditions which are $\bm{r}(\cdot,0)$, $\bm{r}_t(\cdot,0)$, $\{\theta_z^0$,$\theta_y^0$,$\theta_x^0$\} and $\{\Dot{\theta}_z^0$, $\Dot{\theta}_y^0$, $\Dot{\theta}_x^0\}$ (where $\theta_{(\cdot)}^0$ and $\Dot{\theta}_{(\cdot)}^0$ are the initial Tait–Bryan angles of the quadrotor and their time derivatives). 

In the following, the procedure for solving~\eqref{eq:12}-\eqref{eq:15} is introduced. Firstly, the solution domain $(s,t)\in[0,L]\times[0,+\infty)$ is discretized in space. Hence, the cable is discretized into $N$ intervals of length $h_s=L/N$ along $s$ direction, and the motion of these intervals is described by the position, velocity, and acceleration of their nodes (i.e., endpoints): $\bm{r}^i(t)$, $\bm{r}_t^i(t)$ and $\bm{r}_{tt}^i(t)$, in which $i=0,1,\dots,N$. Thus, the deformation of the cable is approximately described by the\ys{central differencing scheme} of nodes:
\begin{equation}
    \bm{r}_s^i\approx \sfrac{(\bm{r}^{i+1}-\bm{r}^{i-1})}{2 h_s}
\label{eq:16}
\end{equation}
\begin{equation}
    \bm{n}^i\approx EA( \left\lVert \bm{r}_s^i \right\rVert _2 -1)\frac{\bm{r}_s^i}{\left\lVert \bm{r}_s^i \right\rVert _2}
\label{eq:17}
\end{equation}
\begin{equation}
\begin{aligned}
    \bm{n}_s^i\approx EA \cdot ({\frac{\bm{r}^{i+1}-2\bm{r}^i+\bm{r}^{i-1}}{h_s^2}}-{\frac{\bm{r}^{i+1}-\bm{r}^i}{h_s\cdot \left\lVert \bm{r}^{i+1}-\bm{r}^i \right\rVert _2 }} \\
    +{\frac{\bm{r}^i-\bm{r}^{i-1}}{h_s\cdot \left\lVert \bm{r}^i-\bm{r}^{i-1} \right\rVert _2 }}).
\end{aligned}
\label{eq:18}
\end{equation}

Equations~\eqref{eq:16}-\eqref{eq:18} discretize the (space) derivatives of~\eqref{eq:12}, where $i=1,2,\dots,N$.
For example, in~\eqref{eq:16}, the (space) derivative of $\bm{r}$ at the ${ (i+1)^{\rm th}}$ node $\bm{r}^{i}$ is discretized via the central difference of its two adjacent nodes: ${ i}^{\rm th}$ node $\bm{r}^{i-1}$ and ${ (i+2)}^{\rm th}$ node $\bm{r}^{i+1}$. 
After substituting~\eqref{eq:16}-\eqref{eq:18} into~\eqref{eq:12},  the motion equation of each node is obtained as in the following,\ys{which is the discretized version of~\eqref{eq:12}:}
\begin{equation}
\begin{aligned}
    \bm{r}_{tt}^i=\frac{E}{\rho_c}\cdot({\frac{\bm{r}^{i+1}-2\bm{r}^i+\bm{r}^{i-1}}{h_s^2}}-{\frac{\bm{r}^{i+1}-\bm{r}^i}{h_s\cdot \left\lVert \bm{r}^{i+1}-\bm{r}^i \right\rVert _2 }} \\
    +{\frac{\bm{r}^i-\bm{r}^{i-1}}{h_s\cdot \left\lVert \bm{r}^i-\bm{r}^{i-1} \right\rVert _2 }})-\frac{\rho_a c_d  \left\lVert \bm{r}_t^i \right\rVert _2 \cdot \bm{r}_t^i}{\rho_c A}-g\bm{{\rm E_z}},
\end{aligned}
\label{eq:20}
\end{equation}

in which $i=1,2,\dots,N$.This~\eqref{eq:20} is similar to the results derived via the series of elastic links model. However, the derivation of the dynamics of the boundary points is different.

The dynamics of the node in position $\bm{r}^0$ coincide with the quadrotor's translational dynamics, which is described by the boundary condition~\eqref{eq:13}. 
%
\ys{However, in~\eqref{eq:13} the variable $\bm{n}(0,t)$ still remains unknown. To calculate it, \eqref{eq:22} is needed, which is the discretized version of~\eqref{eq:7} in the domain $[0,L]$ with boundary condition~\eqref{eq:15} ($\bm{n}(L,t)=[0,0,0]^{\rm T}$):}

\begin{align}
            & -\bm{n}(0,t)-\sum\limits_{i=1}^{N}[\frac{1}{2}\rho_a c_d h_s \cdot (\left\lVert \bm{r}_t^i \right\rVert _2 \cdot \bm{r}_t^i+\left\lVert \bm{r}_t^{i-1} \right\rVert _2 \cdot \bm{r}_t^{i-1})] \nonumber \\ 
             & -\rho_c A L g\bm{{\rm E_z}}=\sum\limits_{i=1}^{N}[\frac{1}{2}\rho_c A h_s \cdot (\bm{r}_{tt}^i+\bm{r}_{tt}^{i-1})].\label{eq:22}
\end{align}
From~\eqref{eq:13} and~\eqref{eq:22}, the motion of the quadrotor’s CoM can be reformulated as follows, which is the discretized version of boundary condition~\eqref{eq:13}:
\begin{equation}
    \begin{aligned}
    \bm{r}_{tt}^0=\frac{1}{m_B+\frac{\rho_c A h_s}{2}}\{-(m_B+\rho_c A L)            
    g\bm{{\rm E_z}}+\bm{{\rm R}}_B \sum\limits_{i=1}^{4} \bm{f}_T^i\\
    -\frac{\rho_c A h_s \bm{r}_{tt}^1}{2}- \sum\limits_{i=2}^{N}[\frac{1}{2}\rho_c A h_s(\bm{r}_{tt}^i+\bm{r}_{tt}^{i-1})]\\
    -\sum\limits_{i=1}^{N}[\frac{1}{2} \rho_a c_d h_s (\left\lVert \bm{r}_t^i \right\rVert _2 \cdot \bm{r}_t^i+\left\lVert \bm{r}_t^{i-1} \right\rVert _2 \cdot \bm{r}_t^{i-1})] \}.\\
    \end{aligned}
\label{eq:23}
\end{equation}
\eqref{eq:23} and~\eqref{eq:14} describe the quadrotor dynamics and provide the boundary condition in $s = 0$. The discretized version of the boundary condition at $s=L$ is expressed as:vc
\begin{equation}
    \bm{r}^{N+1}=\bm{r}^{N-1}+2h_s\frac{\bm{r}^N-\bm{r}^{N-1}}{\left\lVert \bm{r}^N-\bm{r}^{N-1} \right\rVert _2 }.
\label{eq:19}
\end{equation}

The evolution of the motion and deformation of the flexible cable described by ordinary differential equations (ODEs)~\eqref{eq:20} and~\eqref{eq:23} is numerically simulated via fourth-order Runge-Kutta method which is also adopted to numerically solve the attitudinal dynamics of the quadrotor described by~\eqref{eq:14}. In this case, the system described by the state \{$\bm{r}^0,\bm{r}^1,\ldots,\bm{r}^N,\theta _x,\theta _y,\theta _z,\bm{r}_t^0,\bm{r}_t^1,\ldots,\bm{r}_t^N,\Dot{\theta}_x,\Dot{\theta}_y,\Dot{\theta}_z$\} is simulated under the effect of the inputs \{$\omega_1,\omega_2,\omega_3,\omega_4$\}.

\section{Reduced-Order Model}\label{sec:model_reduced}
\revision{The quadrotor-cable system can be numerically simulated using the finite-dimensional model represented by~\eqref{eq:14},~\eqref{eq:20}, and~\eqref{eq:23}. A large $N$ (e.g., $N=100$ for a 1\,m long cable) guarantees a high fidelity of the simulation while a small $N$ does not provide in general acceptable results. A large $N$ however makes it hard to adopt these equations as the plant model for control design due to the large size of the state evolving in ${\rm SE(3)} \times {\rm se(3)} \times \mathbbm{R}^{6N}$.}
\revision{In order to overcome such obstacle,} this work proposes to reduce the order of the model using the POD method for projecting the original infinite-dimension system~\eqref{eq:12}-\eqref{eq:15} on a finite low-dimensional subspace spanned by a limited number of basis functions. POD \revision{captures the most prominent dynamical behaviors of the original system -- thus a better approximation -- with a lower dimension of the state when compared to FDM, therefore it} is commonly adopted as a model order-reducing method in the field of computational fluid dynamics (CFD) \cite{taira2017modal,taira2020modal}. At the best of our knowledge, it is the first time that such a method is applied to this type of robotic systems. 

In the following, the procedure for deriving the Reduced-Order Model (ROM) for the quadrotor-cable system is introduced. Firstly, the orthogonal basis functions, also called \textit{dynamic modes}, of the cable are obtained from simulation data. After that, the original cable dynamics~\eqref{eq:12} is projected on the subspace linearly spanned by those basis functions. Finally, the difference between the original system and the ROM is evaluated.

\subsection{Quadrotor Position Control}
For obtaining a suitable group of basis functions to describe the motion of cable, simulation data are needed. Hence, a hierarchical position tracking controller similar to the one used in~\cite{kendoul2010guidance} has been designed to drive the quadrotor and excite the dynamics of the cable.

The hierarchy of quadrotor position controller consists of two layers: the outer loop is expressed as 
\begin{equation}
    \begin{aligned}
            {\Ddot{\bm{p}}_B}^d=&{\bm{{\rm K}}}_p^r \cdot (\bm{p}_B^r-{\bm{p}_B})+{\bm{{\rm K}}}_d^r \cdot ({\Dot{\bm{p}}_B}^r-{\Dot{\bm{p}}_B})   & {\rm (a)}\\
            \left\lVert \bm{f}_B^d \right\rVert _2=& \left\lVert m \cdot ({\Ddot{\bm{p}}_B}^d+g \bm{{\rm E_z}}) \right\rVert _2  & {\rm (b)} \\
            \left[
                \begin{array}{c}
                    \theta_x^d \\
                    \theta_y^d \\
                    \theta_z^d \\
                \end{array}
            \right]
            =&
            \left[
                \begin{array}{c}
                    \arcsin(\frac{m \cdot [0,1,0]^{{\rm T}} {\Ddot{\bm{p}}_B}^d }{\left\lVert \bm{f}_B^d \right\rVert _2}) \\
                    \arcsin(\frac{m \cdot [1,0,0]^{{\rm T}} {\Ddot{\bm{p}}_B}^d}{\left\lVert \bm{f}_B^d \right\rVert _2 \cdot \cos{\theta_x^d}} ) \\
                    0 \\
                \end{array}
            \right]  & {\rm (c)} \\
                    [\Dot{\theta}_x^d,
                    \Dot{\theta}_y^d,
                  \Dot{\theta}_z^d]^{{\rm T}} 
            =&[0,0,0]^{{\rm T}}  & {\rm (d)}
    \end{aligned}
\label{re_eq:1}
\end{equation}
Eq.~\eqref{re_eq:1} controls the translational motion of the quadrotor CoM by three virtual inputs \{$\left\lVert \bm{f}_B^d \right\rVert _2$,$\theta_x^d$,$\theta_y^d$\} which are the desired magnitude of the total propellers' thrust force, and the desired Tait–Bryan angles along X and Y axis. For calculating these three inputs, the PD law \eqref{re_eq:1}(a) is used to obtain the quadrotor's desired translational acceleration $ {\Ddot{\bm{p}}_B}^d$, where ${\bm{{\rm K}}}_p^r={\rm diag}\{{k_p^r}_x,{k_p^r}_y,{k_p^r}_z\}$ and ${\bm{{\rm K}}}_d^r={\rm diag}\{{k_d^r}_x,{k_d^r}_y,{k_d^r}_z\}$ are the feedback gain matrices of the position loop. After that,  $ {\Ddot{\bm{p}}_B}^d$ is mapped to \{$\left\lVert \bm{f}_B^d \right\rVert _2$,$\theta_x^d$,$\theta_y^d$\} based on~\eqref{re_eq:1}(b,c). Unlike in \cite{kendoul2010guidance}, the mass $m$ used in~\eqref{re_eq:1}(b) is $m=m_B+\rho_c A L$, which is the total mass of the quadrotor and the cable rather than quadrotor mass $m_B$ only. In this work, the desired Tait–Bryan angle along the Z axis $\theta_z^d$ (quadrotor heading) is set to 0, as well as the changing rate of desired Tait–Bryan angles

The inner loop of the quadrotor position controller is  
\begin{equation}
    \left\{
        \begin{aligned}
            \left[
                \begin{array}{c}
                    \Ddot{\theta}_x^d \\
                    \Ddot{\theta}_y^d \\
                    \Ddot{\theta}_z^d \\
                \end{array}
            \right] 
            =&{\bm{{\rm K}}}_p^{\theta} \cdot
            \left[
                \begin{array}{c}
                    {\theta}_x^d-{\theta}_x \\
                    {\theta}_y^d-{\theta}_y \\
                    {\theta}_z^d-{\theta}_z \\
                \end{array}
            \right]
            +{\bm{{\rm K}}}_d^{\theta} \cdot
            \left[
                \begin{array}{c}
                    \Dot{\theta}_x^d-\Dot{\theta}_x \\
                    \Dot{\theta}_y^d-\Dot{\theta}_y \\
                    \Dot{\theta}_z^d-\Dot{\theta}_z \\
                \end{array}
            \right] & {\rm (a)} \\
            \Dot{\bm{\omega}}_B^d
            =&\mathbf{T}^{-1} \cdot ( [\Ddot{\theta}_x^d,\Ddot{\theta}_y^d,\Ddot{\theta}_z^d]^{\rm T}
            -\Dot{\mathbf{T}} \cdot {\bm{\omega}}_B) & {\rm (b)}\\
            \bm{\tau}_B^d
            =&\bm{{\rm J}}_B \Dot{\bm{\omega}}_B^d+\bm{{\rm J}}_B \times {\rm \bm{{\rm J}}}_B\bm{\omega}_B & {\rm (c)}\\
        \end{aligned}
    \right\}
\label{re_eq:2}
\end{equation}
In the control loop in~\eqref{re_eq:2}, the virtual input, namely the desired torque $\bm{\tau}_B^d$, is used to track the orientation references $\{{\theta}_x^d,{\theta}_y^d,{\theta}_z^d,\Dot{\theta}_x^d,\Dot{\theta}_y^d\,\Dot{\theta}_z^d\}$.  ${\bm{{\rm K}}}_p^{\theta}$={\rm diag}$\{{k_p^{\theta}}_x,{k_p^{\theta}}_y,{k_p^{\theta}}_z\}$ and ${\bm{{\rm K}}}_d^{\theta}={\rm diag}\{{k_d^{\theta}}_x,{k_d^{\theta}}_y,{k_d^{\theta}}_z\}$  are the feedback gain matrices of the orientation loop.



\ys{After that, the inputs $\{\left\lVert \bm{f}_B^d \right\rVert _2,\bm{\tau}_B^d\}$ are transformed to the desired rotational velocity of each propeller $\{\omega_1^d,\omega_2^d,\omega_3^d,\omega_4^d\}$ through the usual quadratic model. In this work, we assume the spinning dynamics of the propellers are fast enough, which means the propellers' spinning speed could track their corresponding reference well enough.}

\subsection{Cable Data Collection}

\revision{Assume that the physical parameters of the real cable are identified (see, e.g., Section~\ref{sec:real}), then the data collection for the ROM generation is done by accurately simulating the cable-quadrotor system using the accurate FDM model in closed loop with the control law explained in the previous section.} The FDM-simulated cable is made swinging by letting the quadrotor following a  reference trajectory. For exciting the dynamics of the flexible cable, the reference trajectory of the quadrotor is selected as a sinusoidal sweep signal in three axes of the world frame. After that, the displacement and deformation of the cable are recorded over the discrete sequence of time instants $0, t_s, 2t_s, \ldots, t_f$, where $t_f$ and $t_s$ are the simulation time and the sampling period, respectively. \ys{The number of time intervals is denoted with $S=\frac{t_f}{t_s}$.}  At each time instant the current shape of the cable is spatially sampled by $M+1$ sampling points $\bm{\mathbbm{r}}^j$ with $j=0,1,\dots,M$, where $\bm{\mathbbm{r}}^j \in \{\bm{r}^0,\dots,\bm{r}^N\}$ and it holds that $S \leq M$ (for obtaining all the $M$ dynamic modes of the cable during decomposition process \cite{taira2017modal}). In this work, $\bm{\mathbbm{r}}^j$ is sampled at intervals of length $h_d=L/M$ equally spaced along $s$. 



Especially, the data representing the sampled cable shape evolution over time are stored in the multi-dimensional matrix $\bm{\mathcal{X}}_r \in \mathbbm{R}^{3 \times (M+1) \times (S+1)}$, where the column refers to the displacements of each point along three axes, and the row refers to the projected displacements of sampled points along the cable on each axis, and the tube of the matrix indicates how the displacement varies w.r.t. time:
\begin{equation}
    \left\{
        \begin{aligned}
            {{\bm{\mathcal{X}}}_r}_{:jk} 
            =& \bm{\mathbbm{r}}^{j-1}(k t_s-t_s) & {\rm (a)} \\
            {\bm{\mathcal{X}}_r}_{i:k} 
            =& [{\rm e}_i^{\rm T} \bm{\mathbbm{r}}^{0}(k t_s-t_s), \cdots, {\rm e}_i^{\rm T} \bm{\mathbbm{r}}^{M}(k t_s-t_s) ]^{\rm T} & {\rm (b)} \\
            {\bm{\mathcal{X}}_r}_{ij:} 
            =& [{\rm e}_i^{\rm T} \bm{\mathbbm{r}}^{j-1}(0),{\rm e}_i^{\rm T} \bm{\mathbbm{r}}^{j-1}(t_s), \cdots, {\rm e}_i^{\rm T} \bm{\mathbbm{r}}^{j-1}(t_f) ]^{\rm T} & {\rm (b)} \\
        \end{aligned}
    \right\},
\label{re_eq:4}
\end{equation}
where, ${\rm e}_1 =[1,0,0]^{\rm T}$,
${\rm e}_2 =[0,1,0]^{\rm T}$,
${\rm e}_3 =[0,0,1]^{\rm T}$.



\subsection{Proper Orthogonal Decomposition}

Similar to \cite{taira2017modal}, the displacement field of the cable $\bm{r}(s,t)$ is split into steady component and unsteady component:
\begin{equation}
    \bm{r}(s,t)=\bm{r}_{\rm steady}+\bm{r}_{\rm unsteady}(s,t)
    \label{re:eq:pod:1}
\end{equation}
in which the steady component is obtained as
\begin{equation}
\begin{aligned}
\bm{r}_{\rm steady}=\bm{r}(s,\cdot)&=\bm{r}(0,\cdot)+[0,0,-(1+\frac{\rho_c g L}{E})s+\frac{\rho_c g}{2E}s^2 ]^{\rm T}  \\     
&=\bm{r}(0,\cdot)+
[0,0,r_{\rm steady}(s) ]^{\rm T}
 \end{aligned}
    \label{re:eq:pod:2}
\end{equation}
 by solving the differential equation~\eqref{eq:12}  substituting the equilibrium conditions $\bm{r}_t(s,\cdot)=[0,0,0]^{\rm T}$, $\bm{r}_{tt}(s,\cdot)=[0,0,0]^{\rm T} \, (s \in [0,L])$ and the boundary condition $||\bm{r}_s(L,\cdot)||_2=1$.
After applying POD on the unsteady component of~\eqref{re:eq:pod:1}, the cable's displacement is approximated as:
\begin{equation}
    \bm{r}(s,t) \approx \bm{r}(0,t)+[0,0,{r}_{\rm steady}(s)]^{\rm T}+\sum\limits_{i=1}^{J} 
\left [
\begin{array}{c}
     a^i_x(t) \phi^i_x(s)  \\
     a^i_y(t) \phi^i_y(s)  \\
     a^i_z(t) \phi^i_z(s)  \\
\end{array}
\right],
    \label{re:eq:pod:3}
\end{equation}
where,  $\phi^i_x,\phi^i_y,\phi^i_z$ are the $i^{\rm th}$ modes of $\bm{r}_{\rm unsteady}$ along $x,y,z$ directions, $a^i_x,a^i_z,a^i_z$ are the coefficients of their corresponding modes, and $J$ is the highest order of the POD modes. It should be noted that the POD modes (along the same direction) are orthonormal, which means the inner product between these modes satisfies:
\begin{equation}
    <\phi^i_*,\phi^j_*>=\int_{0}^{L} \phi^i_*(s) \cdot \phi^j_*(s) \,ds=\delta_{ij}=\left\{
        \begin{array}{l}
            1 \, \, {\rm if} \, i=j \\
            0 \, \, {\rm if} \, i \neq j \\
        \end{array}
    \right.
\label{eq:26}
\end{equation}

In this work, there exists a $K \, (K \geq J)$ satisfying that ${r}_{\rm steady}, \phi^i_x, \phi^i_y, \phi^i_z \, (i=1,2,\cdots,J)$ can be represented as the linear combinations of some new POD modes $\phi^i$:
\begin{equation}
    \begin{aligned}
        &\{ {r}_{\rm steady},\phi^1_x,&\cdots, \phi^J_x,\phi^1_y,\cdots, \phi^J_y, \\
        &\phi^1_z,\cdots, \phi^J_z  \} \in {\rm span} \{ \phi^1, \cdots, \phi^K \}. \label{re:eq:pod:4}
    \end{aligned}
\end{equation}
Hence, the cable's displacement field is simplified and decomposed with the POD modes $\phi^i \, (i=1,2,\cdots,K)$:
\begin{equation}
\bm{r}(s,t) \approx \bm{r}(0,t)+\sum\limits_{i=1}^{K} 
([    a^i_x(t),   a^i_y(t) ,a^i_z(t)]^{\rm T} \cdot \phi^i(s))
    \label{re:eq:pod:5}
\end{equation}


To obtain the POD modes $\phi^i$, higher-order singular value decomposition (HOSVD) is adopted to decompose the recorded cable shape data (in the tensor form $\bm{\mathcal{X}} \in \mathbbm{R}^{3 \times (M+1) \times (S+1)}$) as:
\begin{equation}
    \bm{\mathcal{X}= \bm{\mathcal{G}} \times_1 \bm{{\rm U}}_1 \times_2 \bm{{\rm U}}_2 \times_3 \bm{{\rm U}}_3},
\end{equation}
where $\bm{{\mathcal{G}}} \in \mathbbm{R}^{3 \times (M+1) \times (S+1)}$ is the core tensor and $\bm{{\rm U}}_1 \in \mathbbm{R}^{3 \times 3}$, $\bm{{\rm U}}_2 \in \mathbbm{R}^{(M+1) \times (M+1)}$ and $\bm{{\rm U}}_3 \in \mathbbm{R}^{(S+1) \times (S+1)}$ are the factor matrices referring to the principal component in the respective tensor mode. The frontal slices are expressed as:
\begin{equation}
    \bm{{\mathcal{X}}}_{::k}={\bm{{\mathcal{X}}}_r}_{::k}- \bm{\mathbbm{r}}^{\rm head}(kt_s-t_s) \otimes \bm{{\rm J}}_{(M+1),1},
\label{re_eq:5}
\end{equation}
where, $\bm{\mathbbm{r}}^{\rm head} \in \mathbbm{R}^3$ is the recorded position of the cable head tip, and $\bm{{\rm J}}_{(M+1),1} \in \mathbbm{R}^{(M+1) \times 1}$ is all-ones matrix.

The discrete representation of the POD modes of the cable shape displacement is collected in each column of  $\bm{{\rm U}}_2$ which refers to the principal component in the second mode of $\bm{\mathcal{X}}$. $\bm{{\rm U}}_2$ is calculated via the singular value decomposition (SVD) of $\bm{{\mathcal{X}}}_{(2)}$:
\begin{equation}
    \bm{{\mathcal{X}}}_{(2)}=\bm{{\rm U}}_2 \bm{{\rm \Sigma}}_2 {\bm{{\rm V}}_2}^{\rm T},
\label{re_eq:6}
\end{equation}
with matrix $\bm{{\mathcal{X}}}_{(2)} \in \mathbbm{R}^{(M+1) \times 3(S+1)}$ is the second mode unfolding of the tensor $\bm{{\mathcal{X}}}$:  
\begin{equation}
\begin{aligned}
    \bm{{\mathcal{X}}}_{(2)}=&
    \left[ 
        \begin{array}{cccc}
            \bm{\mathbbm{r}}^0(0)^{{\rm T}} & \bm{\mathbbm{r}}^0(t_s)^{{\rm T}}  & \dots & \bm{\mathbbm{r}}^0(t_f)^{{\rm T}}\\
            \bm{\mathbbm{r}}^1(0)^{{\rm T}} & \bm{\mathbbm{r}}^1(t_s)^{{\rm T}}  & \dots & \bm{\mathbbm{r}}^1(t_f)^{{\rm T}}\\
             \vdots & \vdots   & \ddots & \vdots \\
             \bm{\mathbbm{r}}^M(0)^{{\rm T}} & \bm{\mathbbm{r}}^M(t_s)^{{\rm T}}  & \dots & \bm{\mathbbm{r}}^M(t_f)^{{\rm T}}\\
        \end{array}
    \right] \\
    -& \bm{{\rm J}}_{(M+1),1} \otimes [\bm{\mathbbm{r}}^{head}(0)^{{\rm T}},\bm{\mathbbm{r}}^{head}(t_s)^{{\rm T}},\cdots,\bm{\mathbbm{r}}^{head}(t_f)^{{\rm T}}] \\
\end{aligned}
\label{re_eq:7}
\end{equation}
Moreover, it holds that $\bm{{\rm \Phi}}=\bm{{\rm U}}_2=[\bm{{\rm \phi}}_1\,\bm{{\rm \phi}}_2\,\cdots\,\bm{{\rm \phi}}_{M+1}] \in \mathbbm{R}^{(M+1) \times (M+1)}$, $\bm{{\rm \Sigma}}_2 \in \mathbbm{R}^{(M+1) \times 3(S+1)}$ and $\bm{{\rm V}}_2 \in \mathbbm{R}^{3(S+1) \times 3(S+1)}$. $\bm{{\rm \Phi}}$ and $\bm{{\rm V}}_2$ contain the left and right singular vectors of $\bm{\mathcal{X}}$, and matrix $\bm{{\rm \Sigma}}_2$ holds the singular values $(\sigma_1, \sigma_2,\dots,\sigma_{M+1})$.

The ${\rm i^{th}}$ POD mode is represented as the ${\rm i^{th}}$ column vector $\bm{{\rm \phi}}_i$ in $\bm{{\rm \Phi}}$ that satisfies the orthonormal property shown in~\eqref{eq:28}, where $\phi_i^k$ and $\phi_j^k$ are the $k^{\rm th}$ element of vector $\bm{{\rm \phi}}_i$ and $\bm{{\rm \phi}}_j$, respectively: 
\begin{equation}
    <\bm{{\rm \phi}}_i,\bm{{\rm \phi}}_j>={\bm{{\rm \phi}}_i}^{{\rm T}} \bm{{\rm \phi}}_j=\sum\limits_{k=1}^{M} \frac{\phi_i^k}{\sqrt{h_d}} \cdot \frac{\phi_j^k}{\sqrt{h_d}} \cdot h_d=\delta_{ij}.
\label{eq:28}
\end{equation}

The singular values in $\bm{{\rm \Sigma}}_2$ are used to determine the dominating POD modes from $\bm{{\rm \Phi}}$ based on the ratio between energy held by each mode and total energy:
\begin{equation}
    \frac{{E_m}_i}{\sum\limits_{i=1}^{M+1} {E_m}_i}=\frac{\sigma_i^2}{\sum\limits_{i=1}^{M+1} \sigma_i^2}.
\label{eq:29}
\end{equation}

\subsection{Reduced Order Model}
Based on the approximate representation of the cable's displacement field, the motion and deformation of the cable can be described by the evolution of the coefficients $\{a^j_x,a^j_y,a^j_z\} \quad (j=1,2,\dots,K)$ plus the motion of cable's endpoint $\bm{r}(0,t)$. For obtaining the dynamics of these coefficients,~\eqref{re:eq:pod:5} is substituted into~\eqref{eq:12}, and the new expression of the cable dynamics becomes
\begin{equation}
\begin{aligned}
    \bm{n}_s(s,t)-\rho_a c_d \bm{r}_t \cdot |\bm{r}_t|-\rho_c A 
    g\bm{{\rm E_z}}
    =&\sum\limits_{i=1}^{K}\left[ 
        \begin{array}{c}
            \Ddot{a}^i_x(t) \rho_c A \phi^i(s) \\
            \Ddot{a}^i_y(t) \rho_c A \phi^i(s) \\
            \Ddot{a}^i_z(t) \rho_c A \phi^i(s) \\
        \end{array}
    \right]\\
    &+\rho_c A \bm{r}_{tt}(0,t).
\end{aligned}
\label{eq:30}
\end{equation}

To reorganize the expression $\{\Ddot{a}^i_x,\Ddot{a}^i_y,\Ddot{a}^i_z\}$, both sides of~\eqref{eq:30} are multiplied by the POD mode $\phi^i(s)$ using the scalar product definition in~\eqref{eq:26} thus obtaining

\begin{equation}
\begin{aligned}
    \rho_c A \left[ 
        \begin{array}{c}
            \Ddot{a}^i_x\\
            \Ddot{a}^i_y\\
            \Ddot{a}^i_z\\
        \end{array}
    \right]=-[\bm{r}_{tt}(0,t)+g \bm{{\rm E_z}}] \cdot \rho_c A \int_{0}^{L} \phi^i(s) \,ds \\
    +\int_{0}^{L} \bm{n}_s(s,t) \cdot \phi^i(s) \,ds-\rho_a c_d \int_{0}^{L}  \bm{r}_t \cdot \left\lVert \bm{r}_t \right\rVert _2 \cdot \phi^i(s) \,ds.
\end{aligned}
\label{eq:31}
\end{equation}

In this work, the POD modes $\phi^i(s)$ are represented by the discrete form $\bm{{\rm \phi}}_i$. To do so, the  integral operations in~\eqref{eq:31} are approximately calculated through numerical integration as shown in the following~\eqref{eq:32}-\eqref{eq:34}:
\begin{equation}
    \int_{0}^{L} \phi^i(s) \,ds \approx \sum\limits_{j=1}^{M} \phi_i^j \sqrt{h_d}
\label{eq:32}
\end{equation}
\begin{equation}
    \int_{0}^{L} \bm{n}_s(s,t) \cdot \phi^i(s) \,ds \approx \sum\limits_{j=1}^{M} \bm{n}_s(jh_d,t) \cdot \phi_i^j \sqrt{h_d}
\label{eq:33}
\end{equation}

\begin{equation}
    \int_{0}^{L}  \bm{r}_t \cdot  \left\lVert \bm{r}_t \right\rVert _2 \cdot \phi^i(s) \,ds \approx \sum\limits_{j=1}^{M} \bm{r}_t(jh_d,t) \cdot  \left\lVert \bm{r}_t(jh_d,t) \right\rVert _2 \cdot \phi_i^j \sqrt{h_d}
\label{eq:34}
\end{equation}

where 

\begin{equation}
\begin{aligned}
    \bm{n}_s(jh_d,t)\approx & EA \cdot \left[{\tfrac{\bm{r}(jh_d+h_d,t)-2\bm{r}(jh_d,t)+\bm{r}(jh_d-h_d,t)}{h_d^2}}\right.\\
    &-{\frac{\bm{r}(jh_d+h_d,t)-\bm{r}(jh_d,t)}{h_d\cdot \left\lVert \bm{r}(jh_d+h_d,t)-\bm{r}(jh_d,t) \right\rVert _2 }}\\
    &\left.+{\frac{\bm{r}(jh_d,t)-\bm{r}(jh_d-h_d,t)}{h_d\cdot \left\lVert \bm{r}(jh_d,t)-\bm{r}(jh_d-h_d,t) \right\rVert _2 }}\right]
\end{aligned}
\label{eq:35}
\end{equation}

\begin{equation}
        \bm{r}_t(jh_d,t) \approx \sum\limits_{i=1}^{K}\left[ 
        \begin{array}{c}
            \Dot{a}^i_x(t) \phi^i(jh_d) \\
            \Dot{a}^i_y(t) \phi^i(jh_d) \\
            \Dot{a}^i_z(t) \phi^i(jh_d) \\
        \end{array}
    \right]+\bm{r}_t(0,t).
\label{eq:36}
\end{equation}

The dynamics of the POD modes' coefficients are approximated by~\eqref{eq:31}-\eqref{eq:36}. The ROM of the cable is obtained by combining the POD mode dynamics with the dynamics of the cable endpoint $\bm{r}(0,t)$. In this case, the state of the ROM model is $\bm{X}=[{a}^i_x,{a}^i_y,{a}^i_z,\bm{r}(0,t),\Dot{a}^i_x,\Dot{a}^i_y,\Dot{a}^i_z,\bm{r}_t(0,t)]^{{\rm T}} \, (i=1,2,\dots,K)$, and the evolution of the state is formulated as in~\eqref{eq:37}, in which  $\bm{X}^{'}=[\Dot{a}^1_x,\Dot{a}^1_y,\Dot{a}^1_z,\cdots,\Dot{a}^K_x,\Dot{a}^K_y,\Dot{a}^K_z,\bm{r}_t(0,t)]^{{\rm T}}$ and the acceleration of cable's endpoint $\bm{r}(0,t)$, namely of the quadrotor CoM, is treated as the input $\bm{u}$ of the ROM model.%
%

\begin{equation}
\begin{aligned}
    \Dot{\bm{X}}=\bm{{\rm B}}\bm{u}+\bm{f}(\bm{X})=\left[ 
        \begin{array}{c}
            \bm{{\rm O}}_{(3K+3) \times 3}\\
             -\bm{{\rm I}}_3 \cdot \int_{0}^{L} \phi^1(s) \,ds\\
             \vdots\\
             -\bm{{\rm I}}_3 \cdot \int_{0}^{L} \phi^K(s) \,ds 
             \bm{{\rm I}}_3\\
        \end{array}
\right] \bm{r}_{tt}(0,t)\\
    +\left[ 
        \begin{array}{c}
            \bm{X}^{'}\\
            \begin{aligned}
             \frac{1}{\rho_c A} \int_{0}^{L} \bm{n}_s(s,t) \cdot \phi^1(s) \,ds-g \bm{{\rm E_z}} \cdot  \int_{0}^{L} \phi^1(s) \,ds\\ 
            - \frac{\rho_a c_d}{\rho_c A} \int_{0}^{L}  \bm{r}_t \cdot |
        \left\lVert \bm{r}_t \right\rVert _2 \cdot \phi^1(s) \,ds \end{aligned}\\
            \vdots\\
            \begin{aligned}
            \frac{1}{\rho_c A} \int_{0}^{L} \bm{n}_s(s,t) \cdot \phi^K(s) \,ds-g \bm{{\rm E_z}} \cdot  \int_{0}^{L} \phi^K(s) \,ds \\
            - \frac{\rho_a c_d}{\rho_c A} \int_{0}^{L}  \bm{r}_t \cdot \left\lVert \bm{r}_t \right\rVert _2 \cdot \phi^K(s) \,ds\end{aligned}\\
            \bm{{\rm O}}_{3 \times 1}\\
        \end{array}
    \right]\\
\end{aligned}
\label{eq:37}
\end{equation}

\section{Optimal Controller Design}\label{sec:control}
In the previous section, we derived the ROM for the original cable dynamics described by PDEs. In this section, the ROM  is used in a Nonlinear Model Predictive Control (NMPC) framework for realizing quadrotor-cable system regulation control and cable shape trajectory tracking control. 

\subsection{Control Architecture}
\begin{figure*}
\centering
\includegraphics[width=14cm]{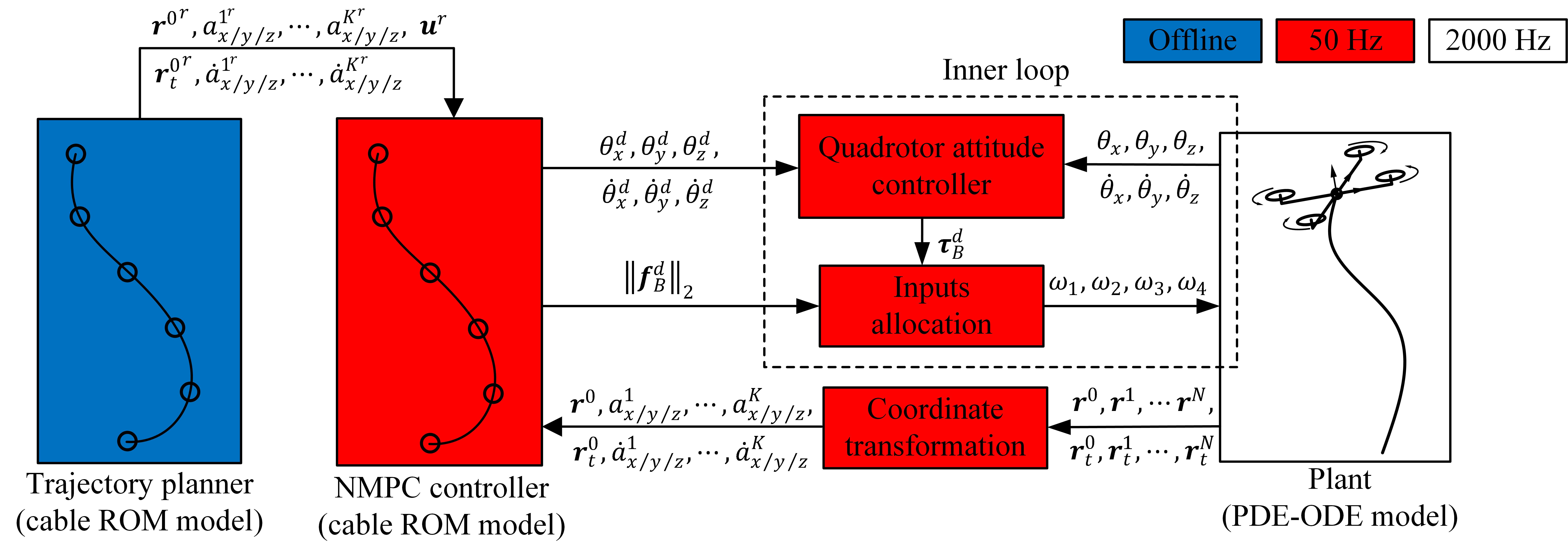}
\caption{\label{fig:2}Control architecture of Quadrotor-cable system (in red, in the center)  connected to the high fidelity PDE-ODE simulator (in white, on the right ) and to the cable trajectory planner (in blue, on the left).}
\end{figure*}

The structure of the cable shape trajectory tracking controller for the quadrotor-cable system is illustrated in Fig.~\ref{fig:2}. The controller is mainly composed of two layers: quadrotor attitude controller (inner loop) and NMPC (outer loop), which receives the reference trajectory from the trajectory planner.

The quadrotor attitude controller is the same as \cite{kendoul2010guidance}; it receives the attitude commands $\{\theta_x^d,\theta_y^d,\theta_z^d, \Dot{\theta}_x^d,\Dot{\theta}_y^d,\Dot{\theta}_z^d\}$ from the outer loop controller and generates desired torque signal $\bm{\tau}_B^d$ for the inputs allocation module.

Additionally, the control architecture involves a coordinate transformation module whose function is to project the high-dimension cable state $\{{\bm{r}^{i}},{{\bm{r}}_t^{i}}\} \,(i=0,1,\dots,N)$ on the subspace spanned by the dominating POD modes. The formulation of the coordinate transformation is expressed in ~\eqref{eq:38} $(i=1,2,\dots,K)$. Eventually, the transformed cable state signals are sent to the NMPC
\begin{equation}
        \begin{array}{c}
            [{{a}^i_x},{{a}^i_y},{{a}^i_z}]^{{\rm T}}=
            \sum\nolimits_{k=1}^{M} ({\bm{r}^{\frac{kN}{M}}}-{\bm{r}^{0}}) \cdot \frac{\phi_i^k}{\sqrt{h_d}} \cdot h_d\\
            {[{\Dot{a}^i_x},{\Dot{a}^i_y},{\Dot{a}^i_z}]}^{{\rm T}}=\sum\nolimits_{k=1}^{M} ({\bm{r}_t^{\frac{kN}{M}}}-{\bm{r}_t^{0}}) \cdot \frac{\phi_i^k}{\sqrt{h_d}} \cdot h_d\\
        \end{array}.
\label{eq:38}
\end{equation}

\subsection{Nonliner Model Predictive Control}
The role of the NMPC module is to solve the optimization control problem (OCP)~\eqref{eq:39} to search for the suitable desired acceleration of the CoM of the quadrotor in the world frame, which is then tracked by the inner loop controller which commands the total thrust $\left\lVert \bm{f}_B^d \right\rVert _2$ and moments $[\theta_x^d,\theta_y^d,\theta_z^d]^{\rm{T}}$.
\begin{align}
    &\underset{ \hat{\bm{u}}_0,\dots,\hat{\bm{u}}_{\frac{t_p}{t_h}-1}}{
   \min}\;
    \sum\limits_{k=1}^{\frac{t_p}{t_h}} [(\hat{\bm{X}}^r_k-\hat{\bm{X}}_k)^{{\rm T}} \bm{{\rm Q}} (\hat{\bm{X}}^r_k-\hat{\bm{X}}_k) + \hat{\bm{u}}_{k-1}^{{\rm T}} \bm{{\rm R}}\hat{\bm{u}}_{k-1}]\notag\\
    &{\rm s.t.}\notag\\
    & \hat{\bm{X}}_0=\bm{X}\label{eq:39}\\
    & {\hat{\bm{X}}}_{k+1}=\bm{F}(\hat{\bm{X}}_k,\hat{\bm{u}}_k)\notag\\
     &\underline{\bm{u}} \leq \hat{\bm{u}}_k \leq \overline{\bm{u}}, \quad k=0,1,\dots,\frac{t_p}{t_h}-1\notag
\end{align}


To minimize the difference between the cable's configuration and its reference trajectory and the control inputs, the optimal control problem of the NMPC is formulated as~\eqref{eq:39}, where $t_p$ is the prediction horizon, $t_h$ is the control time step, and $\hat{\bm{X}}$ and $\hat{\bm{u}}$ are the predicted cable state and predicted control inputs,  respectively. Matrices $\bm{{\rm Q}} \in \mathbbm{R}^{6(K+1)\times6(K+1)}$ and $\bm{{\rm R}} \in \mathbbm{R}^{3 \times 3}$ are the corresponding weight matrices. $\{\hat{\bm{u}}^0,\dots,\hat{\bm{u}}^{\frac{t_p}{t_h}-1}\}$ are govern control inputs at a certain time of the predicted horizon $\{0,t_h,2t_h,\dots,t_p-t_h\}$. Additionally, in order to implement the feedback, the initial state of the predicted model (ROM) $\hat{\bm{X}}_0$ is set equal to the current measured state of the cable $\bm{X}=[{a}^i_x,{a}^i_y,{a}^i_z,\bm{r}(0,t),\Dot{a}^i_x,\Dot{a}^i_y,\Dot{a}^i_z,\bm{r}_t(0,t)]^{{\rm T}} \, (i=1,2,\dots,K)$. The second constraint of the optimization~\eqref{eq:39} means that the evolution of the predicted state follows the ROM dynamics expressed in~\eqref{eq:37}. Indeed, the vector function $\bm{F}$ is the discrete form of the ROM of the cable in~\eqref{eq:37} obtained with the fourth-order Runge-Kutta method. $\underline{\bm{u}}$ and $\overline{\bm{u}}$ are the lower and upper bound of control inputs, respectively.

After each optimization run, the optimal predicted control inputs $\hat{\bm{u}}_0$, i.e., the desired acceleration of the cable head tip ${\bm{r}_{tt}}^d(0,t)$ is transformed  into the desired quadrotor attitude and desired magnitude of propellers’ thrust force using~\eqref{re_eq:1}(b-d), where $\Ddot{\bm{p}}_B^d={\bm{r}_{tt}}^d(0,t)$. Then, These reference signals are processed by the inner attitude loop to generate the spinning speeds of four propellers. 


Any attitude controller can be used for the purpose; in our implementation, we use the attitude controller~\eqref{re_eq:2}, which is also used in the cable data collection process.  

\section{Numerical Experiments}\label{sec:sims}

 In this section, the performance of the reduced-order model and of the proposed controller is verified.
 
 To test the performance of the proposed controller, three scenarios are numerically tested: regulation, cable shape trajectory tracking, and narrow window crossing. In all three scenarios, the controller is based on the ROM but the simulator, purposedly different than the model used by the controller, implements the FDM to solve the PDEs. 
 
 The frequency of the quadrotor attitude controller and inputs allocation module are both set as $50\, \rm{Hz}$. The optimization problem in~\eqref{eq:39} is also solved at a frequency of $50\, \rm{Hz}$, and control step $t_h$ and prediction horizon $t_p$ are set as $0.02 \, {\rm s}$ and $0.4 \, {\rm s}$, respectively.  The number of FDM model nodes is set as $N=100$. Additionally, the dynamics of the FDM model and the ROM (the latter embedded in the NMPC) are simulated via fourth-order Runge-Kutta with fixed step $0.5 \, {\rm ms}$ and with fixed step $20 \, {\rm ms}$, respectively. The physical parameters of the quadrotor-cable system are reported in Table \ref{tab:1}.

 \begin{table}[b]
\begin{center}
\begin{tabular}{||l l l l l l||} 
 \hline
 Item & SI unit & Item & SI unit & Item & SI unit \\ [0.5ex] 
 \hline\hline
 $m_B$ & 0.3 & $\rho_c$ & 1.2732 $\times 10^3$ & $g$ & 9.8 \\
 $E$ &  100000 & $A$ & $7.854 \times 10^{-5} $ & $L$ & 1 \\
 $\rho_a$ &  1.293 & $c_T^i$ & $4 \times 10^{-7} $ & $c_d$ & 0.01 \\
 $c_{\tau}^i$ &  $3 \times 10^{-9}$ & $\bm{{\rm J}}_B$ & \multicolumn{2}{c}{${\rm diag}\{1,1,2\}\times 10^{-6} $}  &  \\
 $\bm{r}_p^1 $ & $[0.15,0,0]^{\rm T}$ & $\bm{r}_p^3 $ & $[-0.15,0,0]^{\rm T}$ &  &  \\
 $\bm{r}_p^2 $ & $[0,0.15,0]^{\rm T}$ & $\bm{r}_p^4 $ & $[0,-0.15,0]^{\rm T}$ &  &  \\[1ex] 
 \hline
\end{tabular}
\caption{Physical parameters of the  quadrotor-cable system.}
\label{tab:1}
\end{center}
\end{table}

%
\subsection{Reduced-Order Model derivation}
Before deriving the ROM, the motion and deformation of the cable were collected in the form of~\eqref{re_eq:4} when the quadrotor followed the reference sinusoidal sweep trajectory in three axes of the world frame. 

\begin{figure}
\centering
\subfloat[POD modes]{%
  \includegraphics[clip,height=3.4cm]{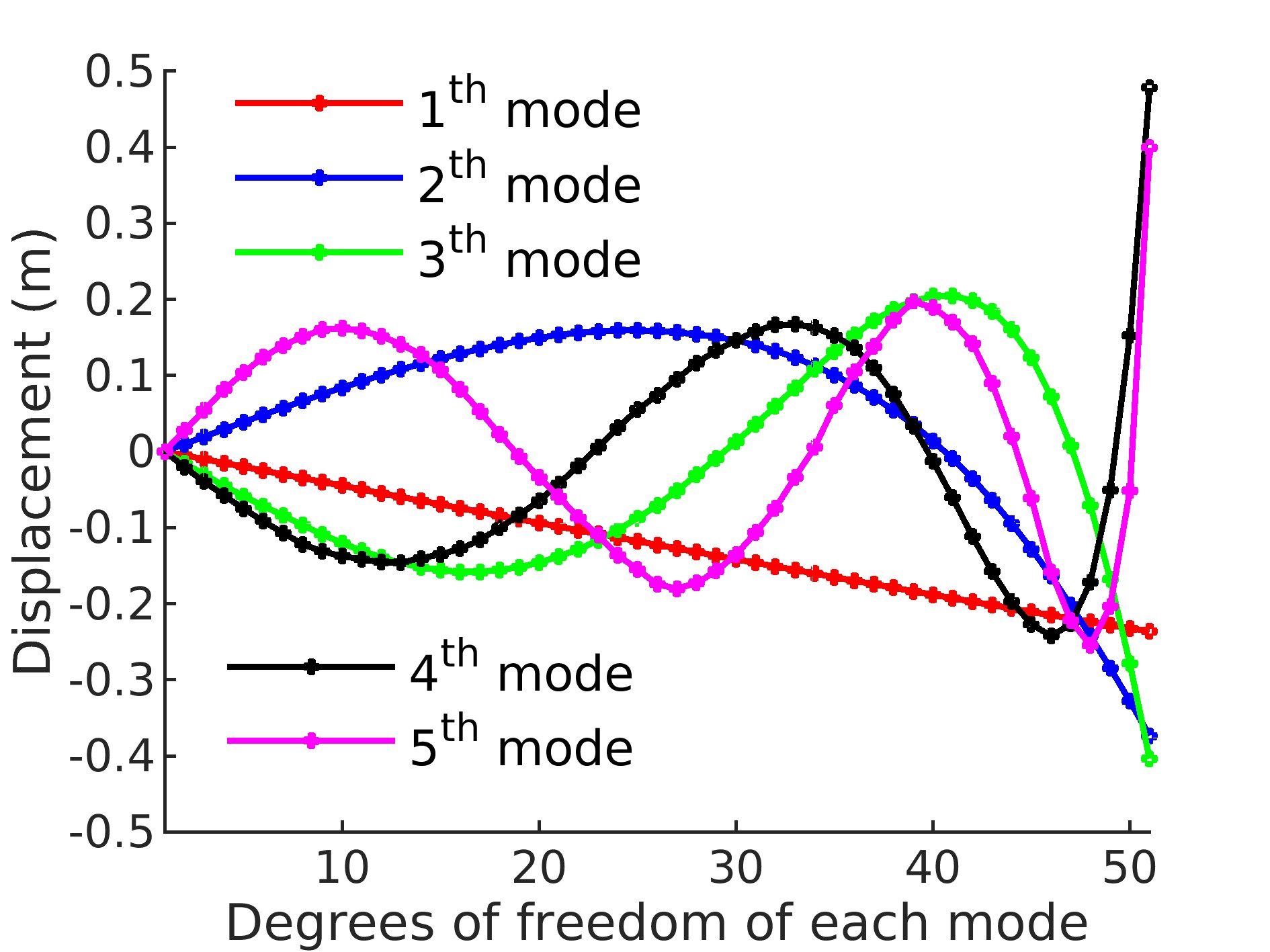}%
}
\subfloat[Energy distribution]{%
  \includegraphics[clip,height=3.4cm]{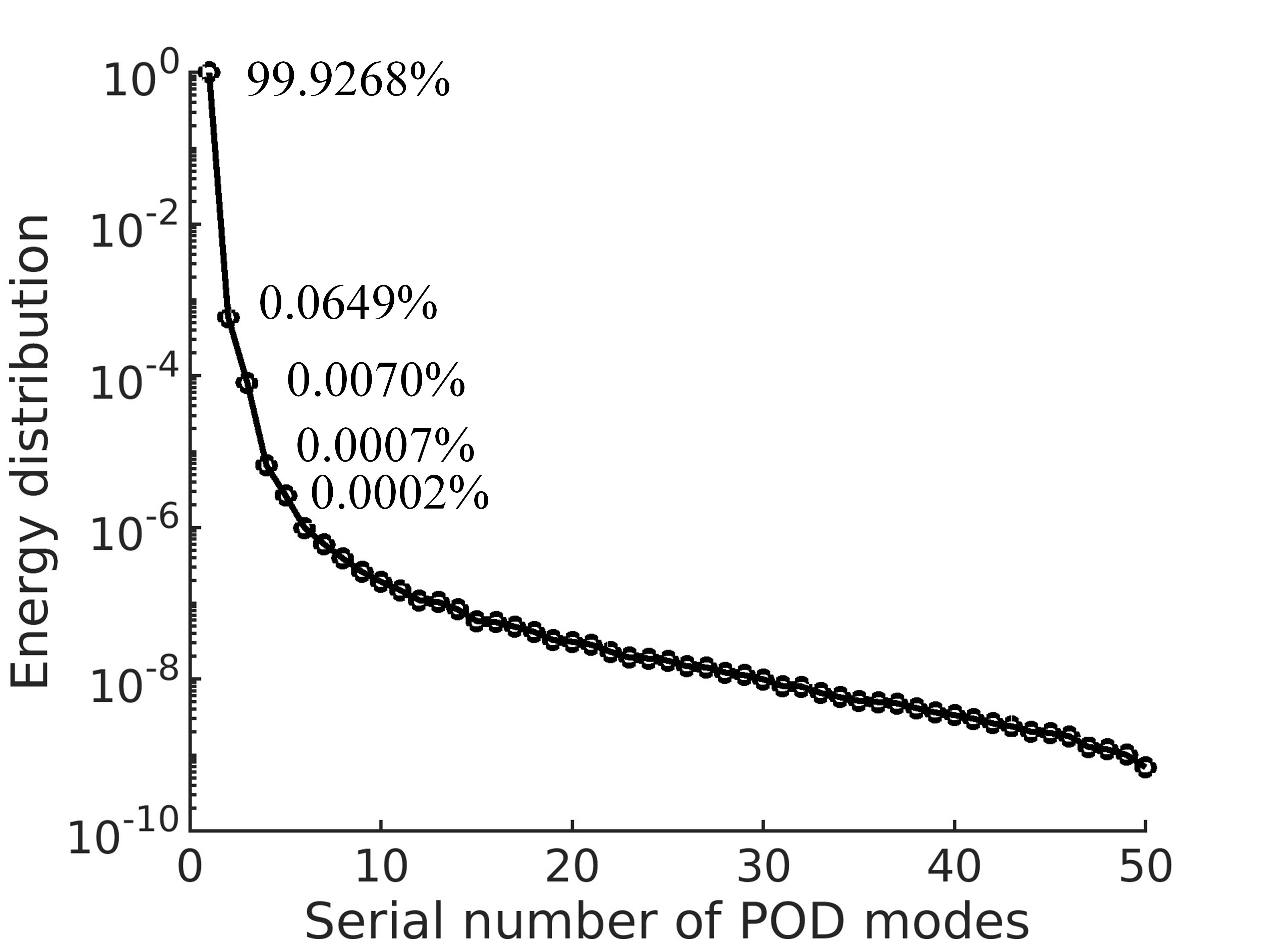}%
}
\caption{Cable POD modes (a) Displacement for first five POD modes of the cable (b) Energy distribution for each POD mode.}
\label{fig:3}
\end{figure}

After that, the positions of material points are recorded in\ys{the tensor~\eqref{re_eq:4}} in which the number of spatial intervals of sampling points $M=50$. After that, the recorded data was decomposed\ys{with~\eqref{re_eq:6}} to obtain the modes of the cable. The first five modes of the cable are illustrated in Figure~\ref{fig:3}(a) for the data gathered through the aforementioned simulation.

According to~\eqref{eq:29}, the energy distribution of each POD mode to the total energy is shown in Figure~\ref{fig:3}(b). It can be found the first POD mode has the major energy. And, the energy distribution of the first three modes is more than $99.998\%$. 

\subsection{Performance of the Reduced-Order Model}
To verify the performance of the ROM derived in Section \ref{sec:model_reduced}, two types of numerical simulation are executed: one uses FDM to compute the cable dynamics (\eqref{eq:16}-\eqref{eq:20}), and the other uses the ROM model (\eqref{eq:32}-\eqref{eq:37}) with different order $K$. For decreasing the complexity of ROM (used in trajectory planning and NMPC procedures), the number of sampling points on the cable which is computed by FDM is set as $M=10$. The initial and boundary conditions of these two simulation types are set equal to each other. The initial configuration of the cable is set equal to $s\mapsto-s\mathbf{E_x}$ (the cable lies horizontally), and the initial velocity of each material point on the cable is put to zero. Hence, the initial state of the FDM is expressed as:
\begin{equation}
        \begin{array}{c}
            {\bm{r}^i}^0=-ih_s\mathbf{E_x} \\
            {\bm{r}_t^i}^0=[0,0,0]^{{\rm T}}\\
        \end{array}
    \quad (i=0,1,\dots,N)
\label{eq:46}
\end{equation}
For the ROM, the cable's initial state is given in~\eqref{eq:47}:
\begin{equation}
        \begin{array}{c}
            [{{a}^i_x}^0,{{a}^i_y}^0,{{a}^i_z}^0]^{{\rm T}}=
            \sum\limits_{k=1}^{M} -kh_d\mathbf{E_x} \cdot \frac{\phi_i^k}{\sqrt{h_d}} \cdot h_d \\
            {[{\Dot{a}_x}^{i^0},{\Dot{a}_y}^{i^0},{\Dot{a}}_z^{i^0}]}^{{\rm T}}=\sum\limits_{k=1}^{M} [0,0,0]^{{\rm T}} \cdot \frac{\phi_i^k}{\sqrt{h_d}} \cdot h_d.\\
        \end{array}
\label{eq:47}
\end{equation}

The cable's boundary condition $\bm{r}(0,t)$ is preset as in~\eqref{eq:48}:
\begin{equation}
    \bm{r}(0,t)=[0,-0.2\cos{\pi \cdot t}+0.2,0]^{\rm T}.
\label{eq:48}
\end{equation}
\begin{figure*}
\centering
\includegraphics[width=18cm]{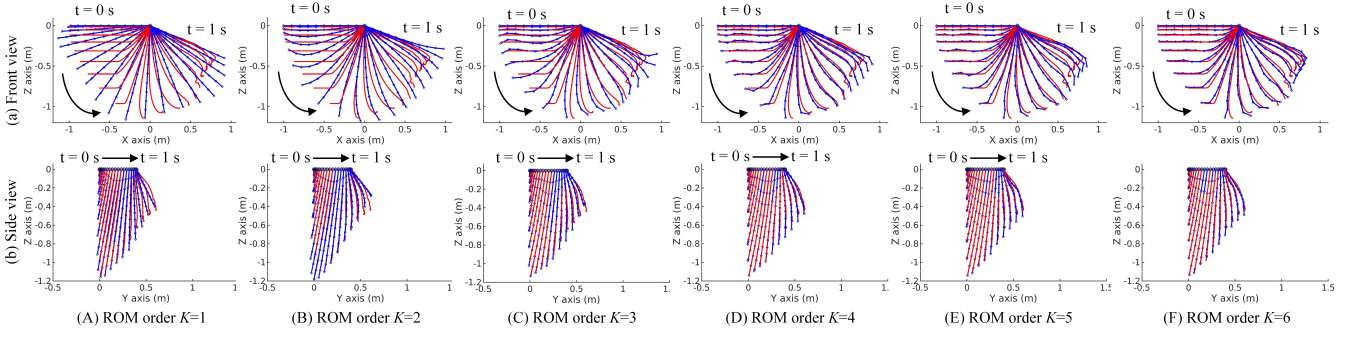}
\caption{\label{fig:5}\ys{Snapshots of the cable simulated using FDM (red) and ROM (blue) with different order $K$.}} 
\end{figure*}
Figure~\ref{fig:5} illustrates the snapshots of the configuration of the cable obtained with the two different models during simulation time $[0 \, {\rm s},1 \, {\rm s}]$. As shown in Figure~\ref{fig:5}(A), it can be appreciated that even with only one mode ($K=1$), the ROM could approximately describe the motion of the cable governed by PDEs. With the ROM order $K$ increasing, the capability of the ROM to describe the deformation of the cable also increases. Additionally, one interesting phenomenon noticeable in Figure~\ref{fig:5} is that the shape of the cable part near the head tip could be captured with low order ROM, while capturing the shape of the cable close to the tail tip requires higher-order ROM. This phenomenon suggests that the cable's tail tip undergoes more complex motion and deformation than the remaining part. 

For quantifying the capability of different-order ROMs to capture the dynamic behavior of the cable governed by PDEs, a metric is induced as:
\begin{equation}
        E_m=\frac{t_{\rm ROM}}{T_{\rm sim}} \sum\limits_{i=1}^{\frac{T_{\rm sim}}{t_{\rm ROM}}} \sum\limits_{k=1}^{M} {\norm{\bm{r}^{\frac{kN}{M}}(i t_{\rm ROM})-\bm{r}_{\rm ROM}(kh_d,i t_{\rm ROM} )}}_2 ,
\label{re_eq:8}
\end{equation}
where, $t_{\rm ROM}= 5 \, {\rm ms}$ is the simulation step of ROMs, $T_{\rm sim} = 5 \, {\rm s}$ is the simulation time duration of the FDM-based model and ROMs, and $\bm{r}_{\rm ROM}$ is the position of material points along the cable's calculated by the ROM.

\begin{figure}
\centering
\includegraphics[width=9cm]{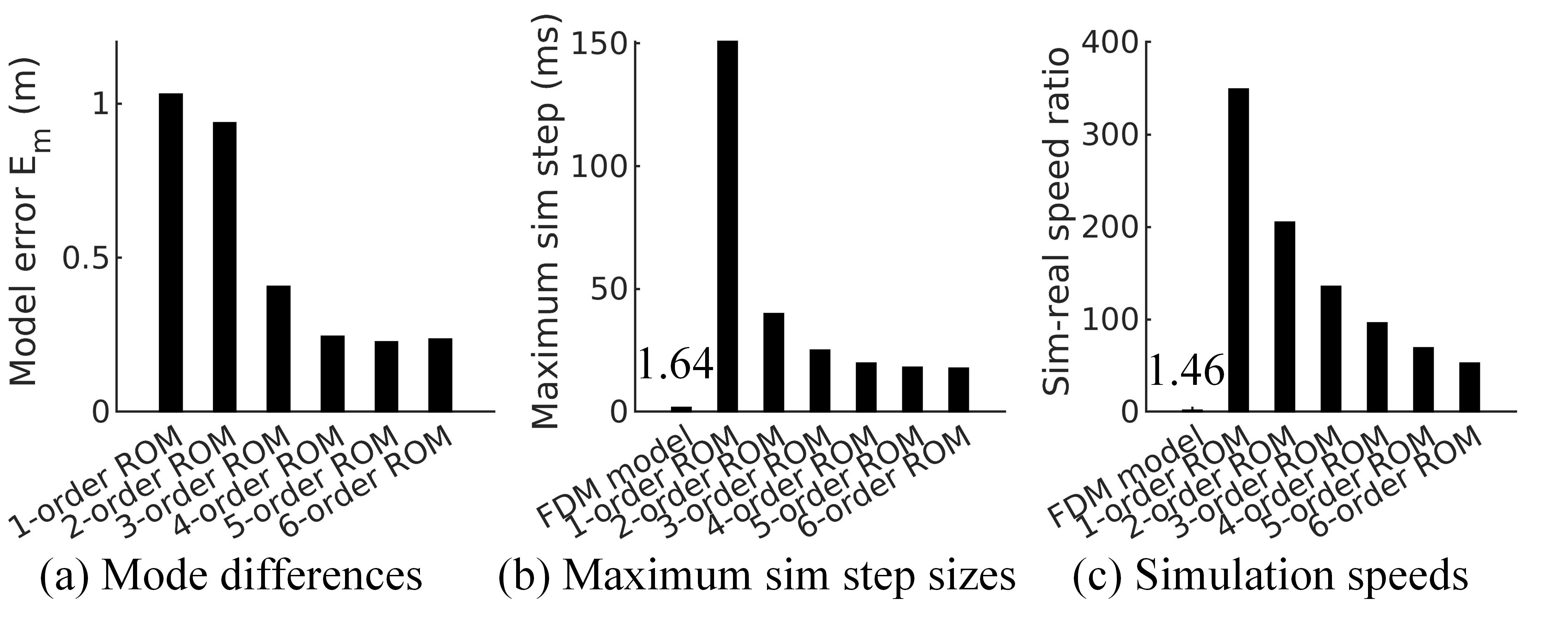}
\caption{\label{re_fig:1}\ys{Comparation between FDM model and ROMs with different truncated order $K$}.}
\end{figure}

\ys{Figure~\ref{re_fig:1}(a) shows the  metric~\eqref{re_eq:8} of different ROMs, where 1-order and 2-order ROMs show higher values of $E_m$. Moreover, with the ROM order $K$ increasing, the ROM could describe the deformation of the cable better (lower $E_m$), while the decrease of $E_m$ is not obvious for $K>3$.}

\ys{The simulation speed and stability are also analyzed for different ROMs. Figure~\ref{re_fig:1}(b) shows the maximum simulation step size of the ROMa and the FDM model. If the simulation time step is bigger than this value, the corresponding model simulation becomes unstable. From Figure~\ref{re_fig:1}(b), it could be found the maximum simulation step size of the FDM model is $1.64 \, {\rm ms}$, which is much lower than the ROM's. The 1-order ROM shows the largest maximum simulation step around $150 \, {\rm ms}$. Additionally, with the ROM order $K$ increasing, the size of the maximum simulation step decreases, while ROMs show a similar maximum simulation step for $K>3$.}

\ys{The simulation speeds of the FDM model and ROMs are shown in~\ref{re_fig:1}(c). In this work, the simulation-real-world speed ratio is adopted to quantify the simulation speed of each model. To obtain this speed metric, each model is simulated for $100 \, {\rm s}$, and each simulation is repeated 50 times. After that, the average time cost is used for calculating the simulation speed. For maximizing the speed of the FDM model, the simulation time step is set as the maximum simulation step size $1.64 \, {\rm ms}$. The simulation time step of all ROMs is set to $t_{\rm ROM}= 5 \, {\rm ms}$, hence this speed is not the maximum simulation speed of ROMs. These simulations are calculated in MATLAB software via the fixed-step fourth-order Runge-Kutta method with CPU Intel® Core™ i5-11400H @ 2.70GHz.}

\ys{From Figure~\ref{re_fig:1}(c), the maximum simulation speed of the FDM model is 1.46 times faster than real-world speed, which is much slower than ROMs. It could be found that the 1-order ROM has the fastest simulation speed (around 350 times faster than real-world speed). And, the simulation speed decreases when the order of ROM increases.}

Compared with modal analysis purpose \cite{taira2017modal}-\cite{taira2020modal}, adopting ROM for model predictive control concerns more factors including model accuracy, model speed, and prediction horizon. The surrogate model used in the predictive controller should be precise enough to predict the behavior of the control plant in the short future. The model is required to be fast (low computational load) so that the predictive controller can find the optimal solution within one control period. Additionally, if the model time step size is larger, with the same prediction steps, the prediction horizon time is longer. In this work, the order of the ROM used in NMPC is $K=3$. According to Figure~\ref{fig:5} and~\ref{re_fig:1}, 3rd-order ROM could capture the motion and deformation of the cable quite well (much better than 1st-order and 2nd-order ROM, and close to higher-order ROM), and it is faster than higher-order ROM and can tolerate a larger simulation time step.

For quantifying the difference between \ys{the FDM model and 3rd-order ROM along time}, two time-dependent \ys{error metrics} expressed in~\eqref{eq:49} are established to measure the shape difference and its rate of change, respectively, where $\bm{r}^{\frac{kN}{M}}(t)$ and $\bm{r}_t^{\frac{kN}{M}}(t)$ are the position and velocity vector of the $(\frac{kN}{M}+1)^{{\rm th}}$ node on the FDM model at time $t$, respectively; $\bm{r}_{\rm ROM}(kh_d,t)$ and ${\bm{r}_{\rm ROM}}_t(kh_d,t)$ are the position and velocity vector of the $(k+1)^{{\rm th}}$ node on the ROM model respectively.

\begin{equation}
\left\{
\begin{array}{c}
    E_1=\sum\limits_{k=1}^{M} {\norm{\bm{r}^{\frac{kN}{M}}(t)-\bm{r}_{\rm ROM}(kh_d,t)}}_2 \\
    E_2=\sum\limits_{k=1}^{M} {\norm{\bm{r}_t^{\frac{kN}{M}}(t)-{\bm{r}_{\rm ROM}}_t(kh_d,t)}}_2  \\
\end{array}
\right\},
\label{eq:49}
\end{equation}

\begin{figure}
\centering
\centering
\includegraphics[width=6cm]{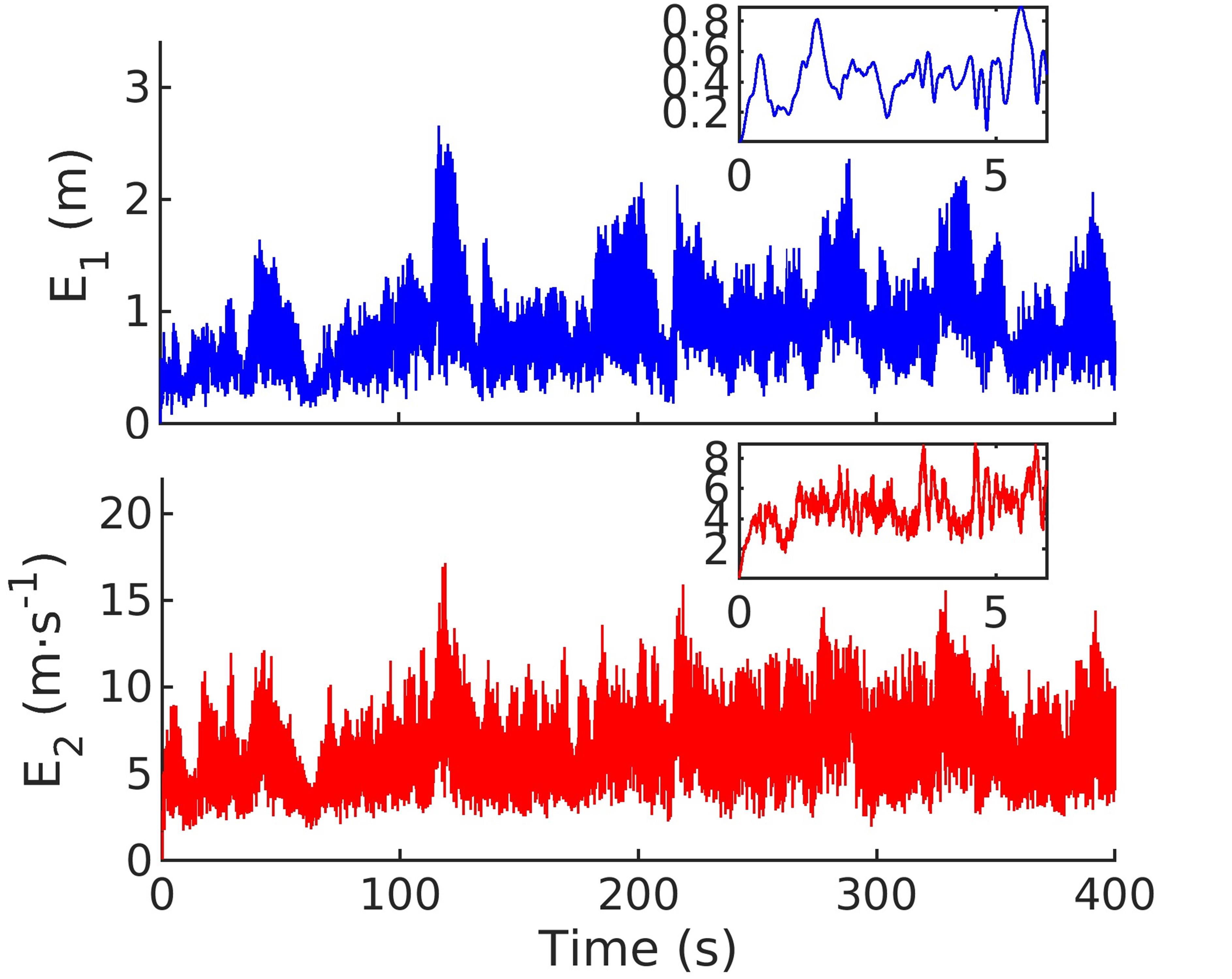}
\caption{Difference between PDM and ROM model.}
\label{fig:6}
\end{figure}

Figure~\ref{fig:6} shows the evolution of two indicators during simulation time $[0 \, {\rm s},400 \, {\rm s}]$. In the zoomed portion on the top right, it can be seen that the shape error and shape change rate error between the two models increases gradually in the first six seconds. After that, the errors remain more or less constant for the following 400 seconds. It can also be noted that the errors exhibit vibrations; those are mainly caused by the phase difference of the cable swing motion.

Figure~\ref{fig:5} and Figure~\ref{fig:6} suggest that, in a short time (less than $4 \, {\rm s}$), the ROM model could approximately describe the deformation and motion of the original PDM model (under the same initial and boundary conditions). With time, the models' differences increase and the correspondence between the two models is gradually lost.

\subsection{Regulation Problem}\label{subs:stab}

In this subsection, we show a simulated scenario \textcolor{black}{with the goal of assessing the capability of the proposed method to regulate the configuration of the quadrotor-cable system to} the static equilibrium, starting from an initial state that is away from the equilibrium state itself. 

To highlight the added value of the proposed control architecture, the same task is executed both with the geometric PID controller~\eqref{re_eq:1}-\eqref{re_eq:2} and the proposed NMPC. The  geometric PID controller only considers the state of the quadrotor, which is tasked to let the quadrotor converge to a certain state or track a desired trajectory. The forces applied on the quadrotor by the cable are regarded as unmodelled external disturbances. Instead, the proposed NMPC, shown in Figure~\ref{fig:2}, considers both the states of the quadrotor and that of the flexible cable.

In this scenario, the initial state of the cable is set as in~\eqref{eq:46}, and the initial attitude state of the quadrotor is set as:
$[\theta^0_x,\theta^0_y,\theta^0_z,\Dot{\theta}^0_x,\Dot{\theta}^0_y,\Dot{\theta}^0_z]^{\rm T}=[0,0,0,0,0,0]^{\rm T}$. The gains of the attitude controller of both NMPC and PID controller are set as $\bm{{\rm K}}^{\theta}_p={\rm diag}\{1500,1500,1500\}$, and $\bm{{\rm K}}^{\theta}_d={\rm diag}\{80,80,80\}$. Additionally, the weight matrices in~\eqref{eq:39} are chosen as $\bm{{\rm Q}}={\rm diag}\{100,100,100,10,10,10,10,10,10,20,20,20,10,10,10,$ $1,1,1,1,1,1,5,5,5\}$, $\bm{{\rm R}}={\rm diag}\{0.001,0.001,0.001\}$. The values of the weight matrix $\bm{{\rm Q}}$ are chosen intuitively, in a way that emphasizes the importance given to the first mode of the cable. 
And, the equilibrium point of the cable is represented by\ys{as~\eqref{re:eq:pod:2}}.

By using~\eqref{eq:38}, the equilibrium state of the cable is transformed to the reference state of the cable's ROM $\bm{X}^r=[{a_x}^{i^r},{{a}_y}^{i^r},{a_z}^{i^r},\bm{r}^r(0,\cdot),{\Dot{a}_x}^{i^r},{\Dot{a}_y}^{i^r},{\Dot{a}_z}^{i^r},\bm{r}_t^r(0,\cdot)]^{{\rm T}} \, (i=1,2,\dots,K)$. In this test, the reference position of the cable upper endpoint is set to  $\bm{r}^r(0,\cdot)=[1,-1,0]^{\rm T}$.

In order to obtain best proportional gains $\bm{{\rm K}}^{r}_p$ and derivative gains $\bm{{\rm K}}^{r}_d$ of the PID controller for the translational motion of the quadrotor, the values of diagonal elements of  $\bm{{\rm K}}^{r}_p$ and $\bm{{\rm K}}^{r}_d$ are searched in the domain of $[0,20]$ to get the minimum value of the following metric
\begin{equation}
f_{e}=\sum\limits_{k=1}^{\frac{t_f}{t_h}} [({\bm{X}}^r-{\bm{X}}_k)^{{\rm T}} \bm{{\rm Q}} ({\bm{X}}^r-{\bm{X}}_k)],
\label{eq:52}
\end{equation}
where, $t_f=15\,s$ is the simulation time, $t_h=0.02\,s$ is the control step time and ${\bm{X}}_k$ is the cable state in ROM sampled at ${k}^{{\rm th}}$ control step.

Differential gradient descent method was implemented to find the best control gains $\bm{{\rm K}}^r_p$ and $\bm{{\rm K}}^r_d$ which minimizes the value of the metric $f_{e}$, and the initial guess of this optimization problem is set as $\bm{{\rm K}}^r_p={\rm diag}\{8,8,8\}$, $\bm{{\rm K}}^r_d={\rm diag}\{4,4,4\}$.  After searching, the gains of the PID controller for the translational errors are chosen as $\bm{{\rm K}}^r_p={\rm diag}\{5.2930,5.6560,10.6966\}$, $\bm{{\rm K}}^r_d={\rm diag}\{2.3990,4.2314,2.5508\}$, while the corresponding minimum metric found is $f_{e}=3697.53$.


\begin{figure*}
\centering
\includegraphics[width=15cm]{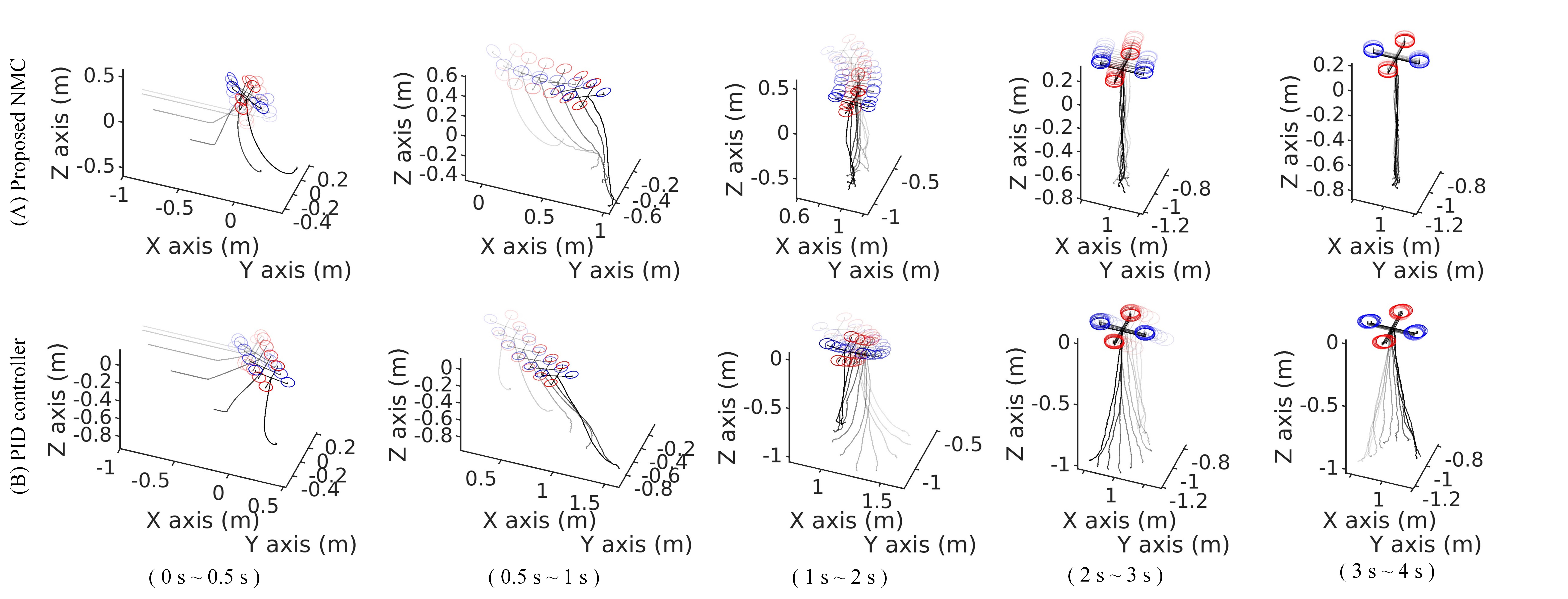}
\caption{\label{fig:7}Snapshot of the quadrotor-cable system controlled with the proposed NMPC (top) and PID (bottom) in the regulation scenario. In each snapshot, multiple consecutive configurations are displayed, where the more transparent ones are older.}
\end{figure*}

\begin{figure*}
\centering
\includegraphics[width=15cm]{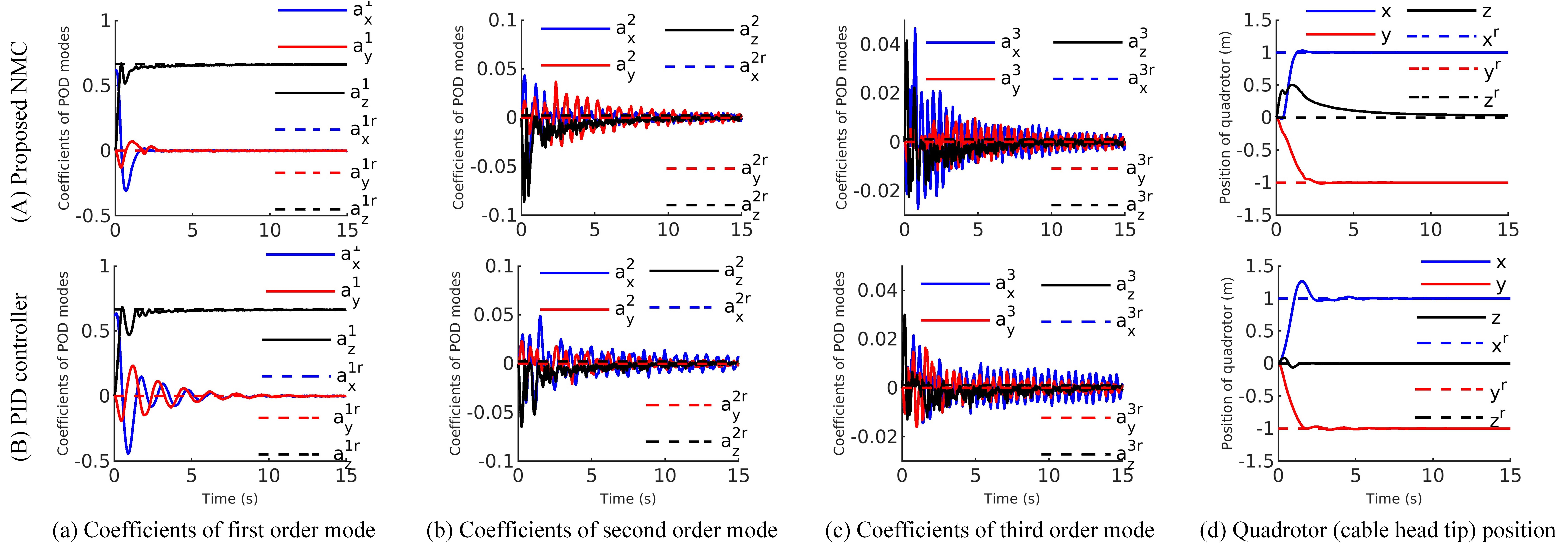}
\caption{\label{fig:8}Evolution of cable state in the regulation scenario using the proposed NMPC (top) and PID (bottom).}
\end{figure*}

Figure~\ref{fig:7} illustrates the two motion sequences of the quadrotor-cable system controlled by the proposed NMPC (Figure  \ref{fig:7}(a)) and the PID (Figure \ref{fig:7} (b)). In Figure \ref{fig:7}(a), after the cable is released from the initial horizontal configuration, the quadrotor tilts to generate translational motion to suppress the swing motion of the cable. It requires $4 \, {\rm s}$ for the quadrotor to approximately suppress the swing of the cable and bring the cable to the target equilibrium as in ~\eqref{re:eq:pod:2}. Instead, in Figure \ref{fig:7}(b), the PID-controlled quadrotor moves to reach its target position failing to effectively suppress the cable oscillations.



Figures \ref{fig:8}(A, B) show the evolution of coefficients of the cable's dominating POD modes $\{{a^i_x}^r,{a^i_y}^r,{a^i_z}^r\} \, (i=1,2,3)$ and the position of the cable endpoint $\bm{r}(0,t)$ with the proposed NMPC and with PID controller, respectively. From Figure~\ref{fig:8}(A-a) and \ref{fig:8}(A-d), we see that the first-order mode and upper endpoint position of the cable converge to the corresponding target state. Inspecting Figure~\ref{fig:8}(B), we see that the PID controller fails to suppress the first-order mode vibrations even after double the time required for the NMPC. Additionally, comparing  \ref{fig:8}(A-b) and \ref{fig:8}(A-c) with \ref{fig:8}(B-b) and \ref{fig:8}(B-c), it is found that the residual vibration amplitudes of the second-order and third-order modes under PID controller are larger, which shows the suppressing ability of NMPC in the three dominating modes of the cable. 

\subsection{Shape Trajectory Tracking}\label{subs:traj_track}

In this second scenario, the quadrotor is required to let the cable track a certain shape trajectory, which includes both the motion and the shape deformation of the cable. \textcolor{black}{The goal is to show the ability of the proposed control method to let the shape of the system track a time-varying configuration. As in the previous section, as a term of comparison, the same task is also executed by a naive PID controller that only forces the quadrotor to track the desired trajectory generated by the ROM and disregards the cable.} Additionally, external disturbances affecting the quadrotor are considered with the goal of assessing the robustness of the control method.  
Finally, uncertainties of different parameters of the cable are induced in numerical experiments to show the effects of each parameter uncertainty on the robustness of the proposed control method. 

The initial state of the system in the second scenario is set as follows: the initial configuration of the cable is set as $s\mapsto-s\bm{{\rm E_z}}$, and the velocity of each material point on the cable is set as zero. The initial attitudinal state of the quadrotor is set as $[\theta^0_x,\theta^0_y,\theta^0_z,\Dot{\theta}^0_x,\Dot{\theta}^0_y,\Dot{\theta}^0_z]^{\rm T}=[0,0,0,0,0,0]^{\rm T}$. The weight matrices of the NMPC, $\bm{{\rm Q}}$ and $\bm{{\rm R}}$ are set as the same as the regulation scenario, and the gains of the PID controller are also set as the same as the previous scenario.

\begin{figure*}
\centering
\includegraphics[width=17cm]{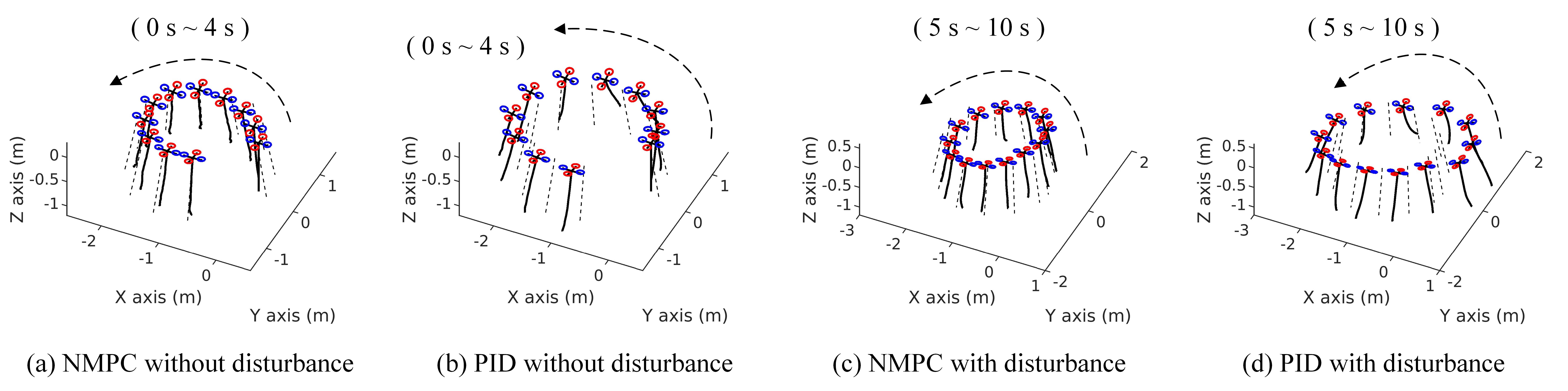}
\caption{\label{fig:10}Snapshot of the quadrotor-cable system controlled with the proposed NMPC or PID controller in the cable shape tracking scenario. In (a) and (b), there is no disturbance effect on the quadrotor-cable system. In (c) and (d), the quadrotor suffers random translational forces in three axes of the world frame at the frequency of 0.5 Hz.}
\end{figure*}

\begin{figure*}
\centering
\includegraphics[width=18cm]{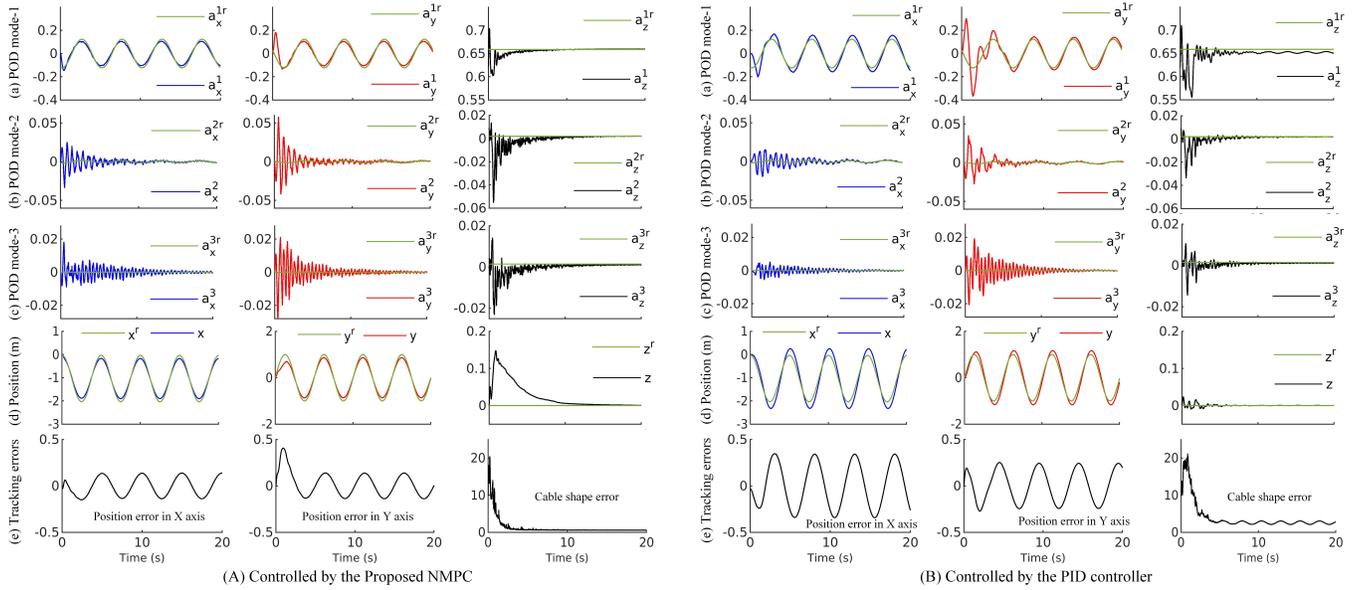}
\caption{\label{fig:11}Evolution of cable state with NMPC (left) and PID controller (right) in the shape tracking scenario without disturbance}
\end{figure*}

The dynamically feasible reference trajectory for the cable in the ROM space is  generated by recording the evolution of the cable in a preliminary simulation where the upper endpoint $\bm{r}(0,t)$ tracks a cycle trajectory expressed on the horizontal plane as
\begin{equation}
    \bm{r}(0,t)=[-1+\cos{\frac{2\pi t}{5}}, \, \sin{\frac{2\pi t}{5}}, \, 0]^{\rm T}.
\label{eq:53}
\end{equation}
After taking the first and second derivatives of ~\eqref{eq:53}, the velocity $\bm{r}_t(0,t)$ and acceleration $\bm{r}_{tt}(0,t)$ of the upper endpoint are obtained. The evolution of the coefficients of dominating modes of the cable is numerically solved with special control inputs $\bm{r}_{tt}(0,t)$ and the initial state of the cable ROM, calculated from~\eqref{eq:38} with the initial configuration of the cable. During this preliminary simulation, the state of the cable converges to a limit cycle whose period is the same as the period of the circular motion of the cable's upper endpoint. Hence, this periodic state trajectory of the ROM is trimmed as the cable's reference shape trajectory shown in the dash lines in Figure \ref{fig:10} and also green lines in Figures \ref{fig:11}(A, B).

Figures \ref{fig:10}(a) and \ref{fig:10}(b) illustrate the motion sequences of the quadrotor-cable system in the cable shape-tracking scenario, while the proposed NMPC and PID controller is implemented, respectively. Comparing the two figures, it is found that with the proposed NMPC, the quadrotor could carry the cable in a neighborhood of the desired state in less time. Conversely, with the PID controller, the cable swings considerably in the first part of the execution, and the cable configuration remains far from the desired one.

Figures \ref{fig:11}(A, B) illustrate the trajectory tracking performance with the proposed NMPC and PID controller, respectively. From Figure~\ref{fig:11}(A), one can see that the coefficients of each POD mode and the position of the cable's upper point converge to the reference trajectory, showing the shape-trajectory-tracking ability of the proposed NMPC. Figure \ref{fig:11}(A-e) shows the tracking errors of the quadrotor for the periodic references in $x$ and $y$ directions, and the total cable shape error computed as 
\begin{equation}
           E_s(t)=[{\bm{X}}^r(t)-{\bm{X}}(t)]^{{\rm T}} \bm{{\rm Q}} [{\bm{X}}^r(t)-{\bm{X}}(t)]
\label{eq:54}
\end{equation}
where the reference state is indicated as $\bm{X}^r(t)$ and the actual one is $\bm{X}(t)$. In~\eqref{eq:54}, the weight matrix $\bm{{\rm Q}}$ is the same as previously defined.

Comparing the results in Figures \ref{fig:11}(A) with those in Figures \ref{fig:11}(B),  it is found that the PID controller requires more time to let the first-order mode coefficients converge, and it shows less accuracy in their tracking. In addition, Figure~\ref{fig:11}(A-e) and \ref{fig:11}(B-e) show that the PID controller performs worse than NMPC in the quadrotor tracking and it gives a higher overall shape-tracking error of the cable at convergence. From \ref{fig:11}(B-d), a non-negligible phase delay can be noticed between reference trajectory and actual state in $x$ and $y$ direction with PID controller. 


\subsubsection{External disturbances}

In this subsection, the performances of the two controllers are tested under disturbance conditions. The quadrotor is assumed to suffer external translational forces during the same task execution as in the first part of Section \ref{subs:traj_track}. Hence, the boundary conditions of the cable (translational dynamics of the quadrotor) in~\eqref{eq:13} are replaced by:
\begin{equation}
m_B \bm{r}_{tt}(0,t) =-m_B g \bm{{\rm E_z}}+\mathbf{R}_B \sum_{i=1}^{4} \bm{f}_T^i + \bm{n}(0,t) + \bm{d}(t)
\label{eq:55}
\end{equation}
where $\bm{d} \in \mathbbm{R}^{3}$ is the external disturbance force along three axes of the world frame. The three components of $\bm{d}$ are assumed to be random\ys{rectangular} wave signals whose amplitude is bounded in the set $[-1,1]$ and frequency set as 0.5 Hz.

\begin{figure*}
\centering
\includegraphics[width=18cm]{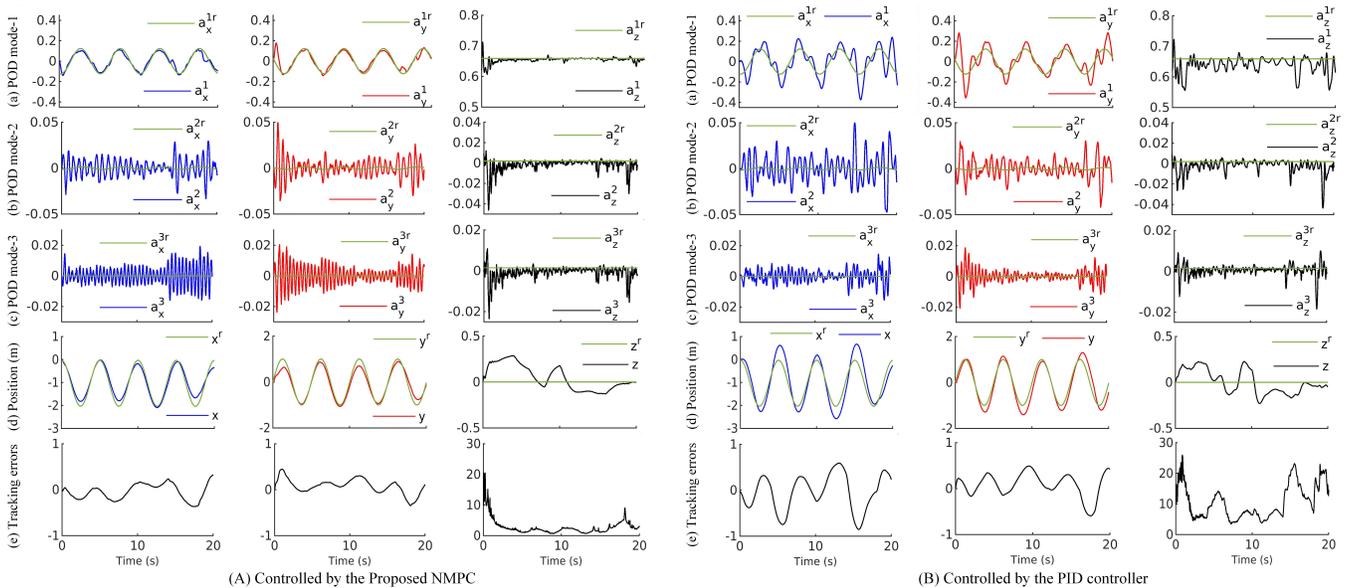}
\caption{\label{fig:13}Evolution of cable state with NMPC (left) and PID controller (right) in the shape tracking scenario with disturbance}
\end{figure*}



Figures \ref{fig:10}(c) and \ref{fig:10}(d) illustrate the trajectory tracking motion sequences of the quadrotor-cable system affected by external disturbance, while the proposed NMPC and PID controller are implemented, respectively. In both cases, the random square wave signals $\bm{d}$ are the same. Comparing Figures \ref{fig:10}(c) and \ref{fig:10}(d), it can be seen that the cable shape tracking performance with the proposed NMPC is better than with the PID controller. 
Indeed, when the PID is used, the cable undergoes a clear swinging motion due to unsuppressed external disturbances. 

The results of the task execution under disturbances are reported in  Figure \ref{fig:13}(A, B) for the proposed NMPC and the PID controller, respectively. From Figures~\ref{fig:13}(A-a) and Figure~\ref{fig:13}(A-d), one can see that the coefficients of the first POD mode and the position of the quadrotor could follow their corresponding references quite well. In Figures~\ref{fig:13}(A-b) and figures~\ref{fig:13}(A-c), the coefficients of the cable's second- and third-order mode remain in the neighborhood of the desired reference.

Comparing the results in Figures \ref{fig:13}(A) with those in Figures \ref{fig:13}(B), the tracking performance obtained with the PID controller emerges to be especially worse in the first-order mode than with the proposed NMPC. The errors in tracking the quadrotor position and the overall cable shape are also higher in the former case.  
These tests show that with the proposed NMPC, the quadrotor was able to track the reference cable shape. \textcolor{black}{It is reasonable to expect analogous performance for other similar reference trajectories.}

\subsubsection{Model parameter uncertainty}

In this section, the effects of different model parameter uncertainties on the cable shape trajectory tracking performance are also analyzed. To do so, we use the metric $E_t$ as in~\eqref{eq:56} to evaluate the cable shape tracking error, where $t_s=20 \, {\rm s}$ and $t_h=0.02 \, {\rm s}$ are the duration of the simulation and control time step, respectively; ${\bm{X}}^r_k$ and ${\bm{X}}_k$ is the reference state and actual state of the cable in ROM sampled at ${k}^{{\rm th}}$ control step.

\begin{equation}
           E_t={\frac{t_h}{t_s}} \sum\limits_{k=1}^{\frac{t_s}{t_h}} [({\bm{X}}^r_k-{\bm{X}}_k)^{{\rm T}} \bm{{\rm Q}} ({\bm{X}}^r_k-{\bm{X}}_k)]
\label{eq:56}
\end{equation} 

%
\begin{figure*}[t]
\centering
\includegraphics[width=14cm]{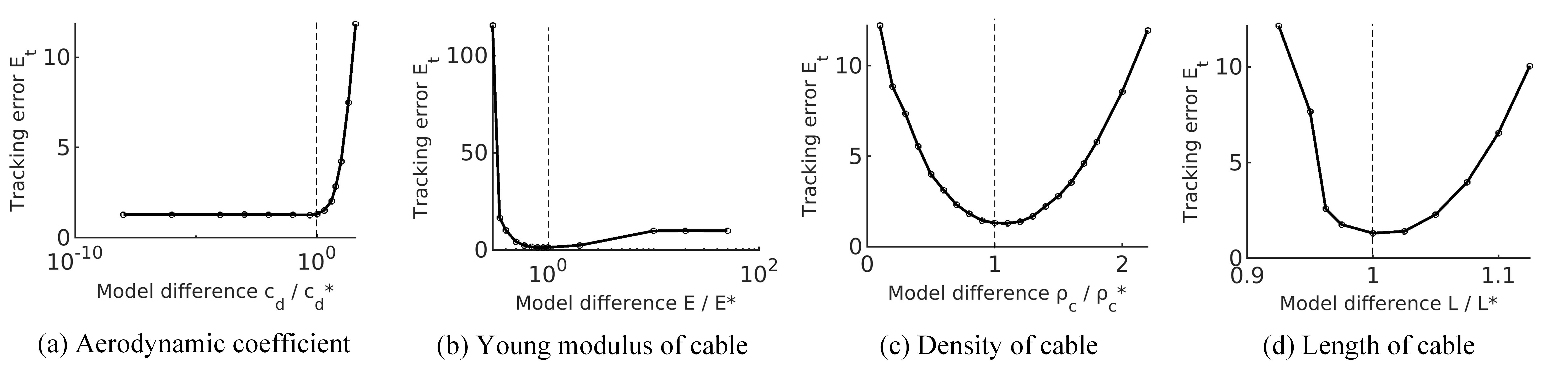}
\caption{\label{fig:15} Dynamic shape tracking errors with different model parameter uncertainties with proposed NMPC}
\end{figure*}
\begin{figure*}[t]
\centering
\includegraphics[width=16cm]{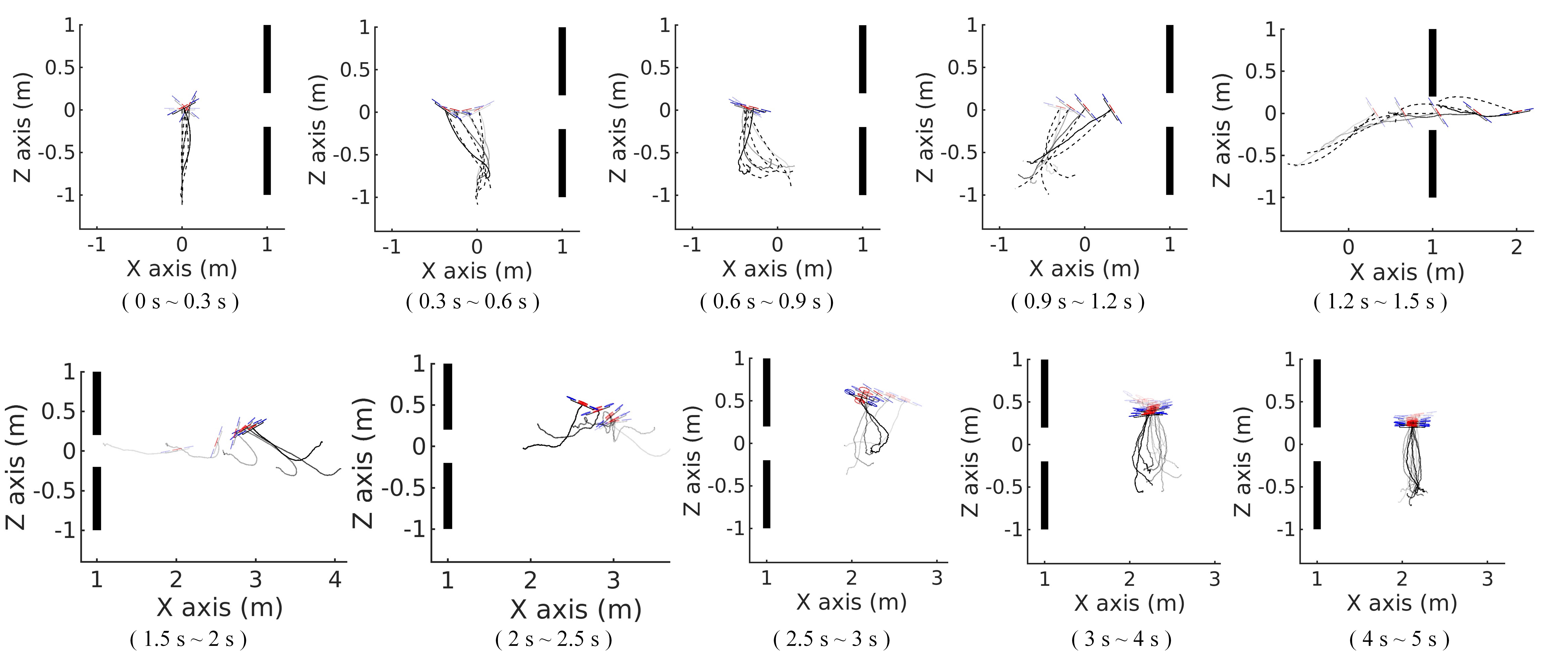}
\caption{\label{fig:16}Snapshot of quadrotor-cable system during window crossing}
\end{figure*}
\begin{figure}[t]
\centering
\includegraphics[width=8.5cm]{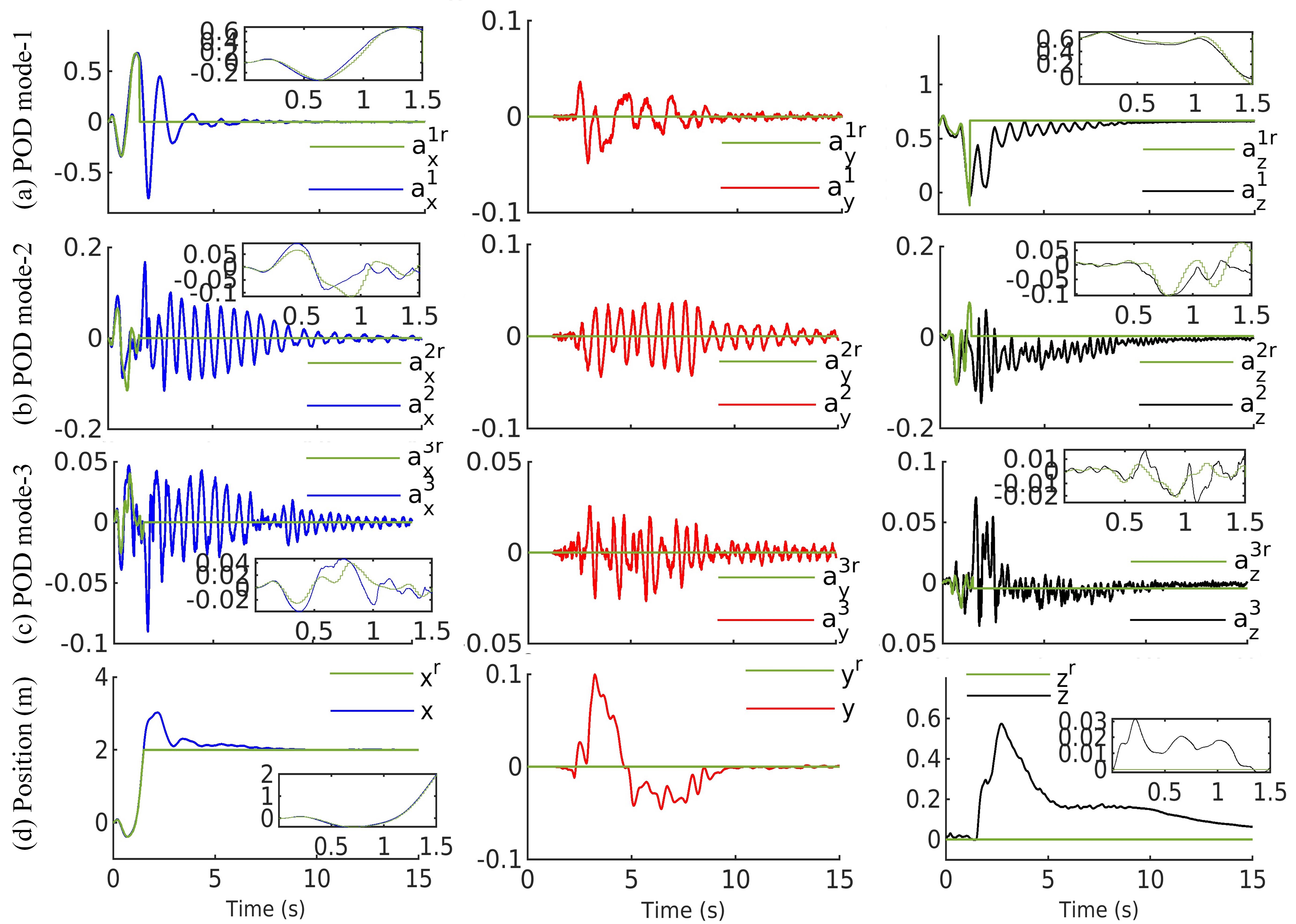}
\caption{\label{fig:17}Evolution of cable state during window crossing.} 
\end{figure}

Figure~\ref{fig:15} shows the tracking errors with different uncertainties on different parameters: the cable's aerial drag coefficient $c_d$, its Young modulus $E$, density $\rho_c$, and length $L$. We found that the effects of a positive or a negative uncertainty on  $c_d$ or $E$ are asymmetric. Especially,  when the real value of the aerodynamic coefficient $c_d$ is greater than the known one, $c_d^\ast$, the tracking error $E_t$ increases sharply. On the contrary, if $c_d$ is lower than $c_d^\ast$, the tracking error $E_t$ remains almost stationary. Interestingly, an actual drag coefficient lower than the one used by the controller seems to be negligible for the task execution. For the Young modulus, the situation is reversed: if the $E$ is higher than its nominal value, the tracking error $E_t$ slightly and slowly increases and then maintains a constant value; when $E$ decreases, on the other hand, the tracking error sharply increases. This can be intuitively explained: when the Young modulus of the cable is high, the cable will basically not change its length under the effect of the forces involved in the manipulation task. 
Instead, when $E$ is reduced, the cable becomes more and more compliant and elongates under the effect of gravity and other forces, causing an increase in the shape trajectory error.  

The effects of uncertainties on the cable density $\rho_c$ and length $L$ are nearly symmetric. When changing the value of $\rho_c$ and $L$ w.r.t. to the value used in the controller, the tracking error $E_t$ increases. Additionally, the tracking error $E_t$ is more sensitive to the uncertainty of cable length: it was found that a $\pm 10\% $ error in $L$ leads to an error comparable to what results from a $100 \%$ error in the cable density.

\subsection{Narrow Window Crossing}\label{subs:window_crossing}

In this scenario, the quadrotor is required to carry the cable through a narrow window, as schematically represented in Figure~\ref{fig:16}. In the window crossing, the cable and the quadrotor itself should not collide with the boundary of the window area. In this scenario, the window is a vertical square opening  $0.4 \, {\rm m}$ wide and its center is placed in the point $[1,0,0]^{{\rm T}}$ in the world frame. The collision constraint is \begin{equation}
    \left\{
        \begin{array}{c}
            -0.2 < {r}^y(s,\cdot) < 0.2 \\
            -0.2 < {r}^z(s,\cdot) < 0.2 \\
        \end{array}
    \right\}
    \quad \text{if}\quad \, {r}^x(s,\cdot)=1,
\label{eq:57}
\end{equation} and the initial state of the system is the same as in Sec. \ref{subs:traj_track}. The target configuration of the cable is the equilibrium state expressed as~\eqref{re:eq:pod:2} where the position of the endpoint is set as $\bm{r}(0,\cdot)=[2,0,0]^{\rm T}$.

An offline trajectory planner is designed to generate a feasible shape trajectory for the cable that satisfies the constraint. After that, the proposed NMPC is used for the quadrotor to let the cable track the designed cable shape reference trajectory. Finally, the NMPC is switched to stabilization-control mode, the same as in the first scenario in Sec. \ref{subs:stab}, to let the cable converge to the final target equilibrium configuration.

The trajectory planner is implemented as follows. First, only the motion of the cable upper endpoint in the $x$ direction is considered. Then, the Fourier series expressed in~\eqref{eq:58} is used for representing the translational motion of the quadrotor (cable upper endpoint). In~\eqref{eq:58}, $\{a_0,\dots,a_{K_r},b_1,\dots,b_{K_r}\}$ are the coefficient of the Fourier series, and $t_r=1.5 \, {\rm s}$ is the time duration of the planned trajectory for crossing the narrow window. After that, by taking the first and second derivatives of~\eqref{eq:58}, the velocity $\bm{r}_t(0,t)$ and acceleration $\bm{r}_{tt}(0,t)$ of the cable endpoint are obtained.
\begin{equation}
    \bm{r}(0,t)=[a_0 + \sum\limits_{i=1}^{K_r} (a_i \sin{\frac{i \pi t}{t_r}} + b_i \cos{\frac{i \pi t}{t_r}}),0,0]^{\rm T}
\label{eq:58}
\end{equation}

The reference shape trajectory $\bm{X}^r(t)$ of the cable is obtained by solving the following feasibility problem~\eqref{eq:59}, where ${r}^x$ is the first component of~\eqref{eq:58}, $\bm{X}$ is the initial state of the cable, $\underline{r_{tt}^x}=-20$ and $\overline{r_{tt}^x}=20$ are the lower and upper boundary of the quadrotor acceleration in the $x$ direction, respectively. Then, the particle swarm optimization (PSO) algorithm is used to find the feasible solution.
\begin{equation}
 \begin{aligned}
    \begin{array}{c}
    \min \\
    a_0,\dots,a_K,\\
    b_1,\dots,b_K
    \end{array} & 
    \begin{array}{c}
    0\\
    \\
    \end{array}\\
    {\rm s.t.} \quad
    & {\bm{X}}^r(0)=\bm{X}\\
    & 2 \leq r^x(0,t_p) \leq 3\\
    & \underline{r_{tt}^x} \leq {r}_{tt}^x(0,t) \leq \overline{r_{tt}^x}\\
    & \Dot{\bm{X}}^r(t)=\bm{f}({\bm{X}^r}(t))+\bm{{\rm B}}{\bm{r}_{tt}}(0,t)\\
    & -0.2 < {r}^y(s,\cdot) < 0.2 \quad {\rm if} \, {r}^x(s,\cdot)=1\\
    & -0.2 < {r}^z(s,\cdot) < 0.2 \quad {\rm if} \, {r}^x(s,\cdot)=1,\\
 \end{aligned}
\label{eq:59}
\end{equation}

Figure~\ref{fig:16} illustrates the motion sequence of the quadrotor-cable system passing through the narrow window area. The first five snapshots $(0 \, {\rm s} \sim 1.5 \, {\rm s})$ show the motion of the quadrotor-cable system and also the reference motion of the cable (dashed line) in the cable shape trajectory tracking stage. In this first stage, the control gains of the proposed NMPC are set as $\bm{{\rm Q}}={\rm diag}\{10,10,10,10,10,10,10,10,10,200,200,200,10,10,10,$ $1,1,1,1,1,1,20,20,20\}$ (while other gains are the same as in 
 Sec. \ref{subs:traj_track}) to emphasize the translational motion error of the quadrotor. With the proposed NMPC, the quadrotor can carry the cable to approximately track the planned reference shape trajectory. After $1.5 \, {\rm s}$, the NMPC switches to regulation mode (the control gains remain the same as the Sec. \ref{subs:stab}), and it requires around $10 \, {\rm s}$ to let the cable converge to the target configuration.

Figure~\ref{fig:17} illustrates the evolution of the coefficients of each POD mode of the cable and the position of the cable's upper point. In the first stage, from $0 \, {\rm s}$ to $1.5 \, {\rm s}$, the coefficients of each POD mode of the cable and the position of the cable’s upper point follow their corresponding references, after $1.5 \, {\rm s}$, the coefficients of each POD mode of the cable and the position of the cable's upper point converge to their references, which also verifies the cable shape trajectory tracking ability of the proposed controller.

\section{Real-world Experiments}\label{sec:real}

In this section, the performance of the proposed PDE model and reduced-order model is verified via real-world experiments, and, the proposed optimal cable shape controller is implemented in real-time, and tested in regulation and tracking scenarios.

\subsection{Experiment setup}

The robotic platform used in real-world experiments is shown in Figure~\ref{fig:1}(a), in which one tip of a 1-meter cable is attached to the bottom of a quadrotor. The quadrotor is designed and developed based on a 10\,cm FPV drone frame, mounted with 4 brushless DC motors (EMAX ECO 1407 motor-2800 KV) and 4 propellers (3525, 3-blade). The quadrotor powered with a 3s LiPO battery (650MAH, 75C) can generate a maximum thrust of around $6 \, {\rm N}$, while its weight (with battery) is 233 grams. 


The flight controller board mounted on the quadrotor is Crazyflie 2.0, while its built-in Mellinger attitude controller (running in $250 \, {\rm Hz}$) \cite{Mellinger2011Minimum} takes the role of the inner loop (quadrotor attitude controller and inputs allocation) of the whole control structure shown in Fig.~\ref{fig:2}.

The software framework implemented for the robotic system is based on Crazyswarm 
\cite{preiss2017crazyswarm}, which is used for the radio communication between the offboard computer and the quadrotor and the WiFi communication between the offboard computer and the motion tracking system (Optitrack, NaturalPoint, Inc.).

To measure the shape of the cable, reflective tapes are attached to the cable, and the arc length between each two close tapes is $h_{\rm exp} = 10 \, {\rm cm}$ along the untensioned cable. Hence, the motion capture system could obtain the point cloud of the cable at $100 \, {\rm Hz}$. The cable point cloud is processed according to~\eqref{eq:38} to get the cable POD mode coefficients. After differentiating the cable POD mode coefficient signals, the change rate of the cable shape could also be obtained.

\begin{figure}[t]
\centering
\includegraphics[width=9cm]{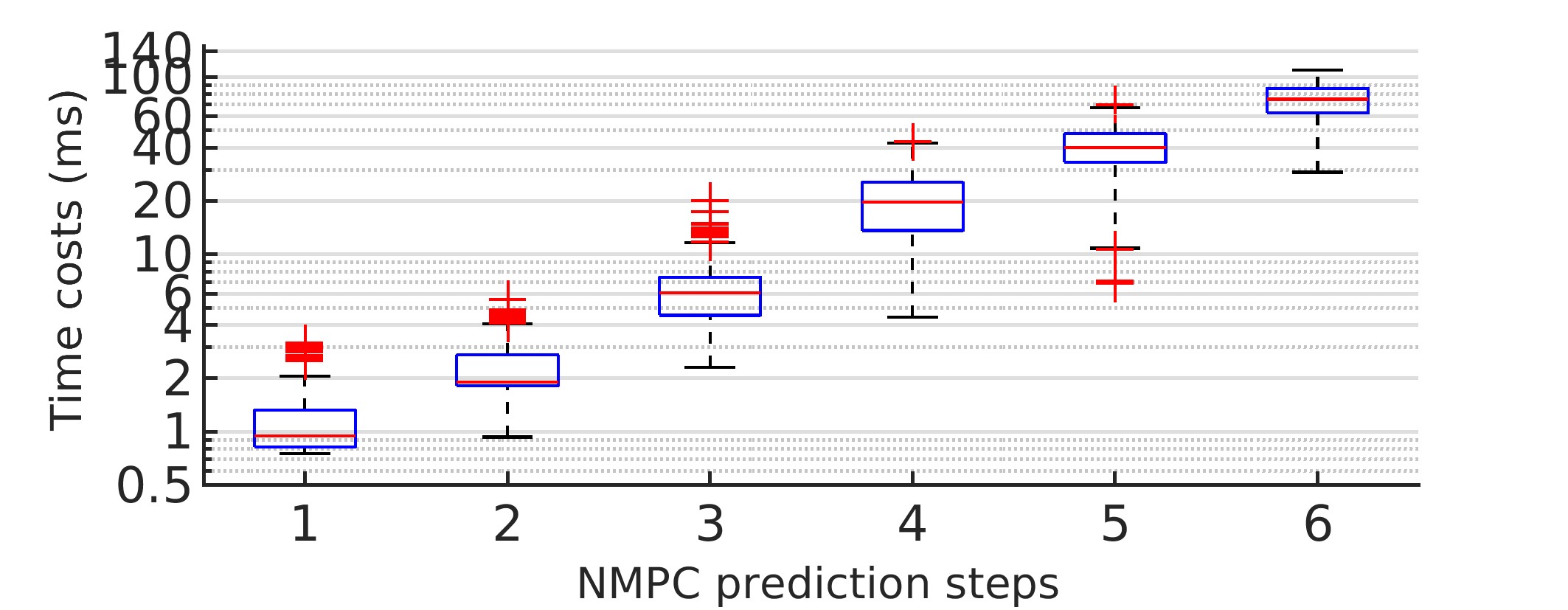}
\caption{\label{re_fig:2}\ys{NMPC Computational load}} 
\end{figure}

For the optimization control problem~\eqref{eq:39} of NMPC, a non-convex optimization framework OpEn~\cite{Sopasakisn2020open} has been implemented in real-time to solve the OCP in C++. The NMPC runs at $100 \, {\rm Hz}$ and the predicted horizon is three steps (each step is $10 \, {\rm ms}$) via the fixed-step fourth-order Runge-Kutta method. This NMPC runs offboard (commercial laptop with  CPU Intel®
Core™ i5-11400H @ 2.70GHz). The average time for solving OCP~\eqref{eq:39} is around $6.14 \, {\rm ms}$. Figure~\ref{re_fig:2} shows the computational time of NMPC with different prediction steps. It is found that an increased prediction step leads to an increase in the computational cost, too. When the prediction steps are 3, the time for solving the OCP of the NMPC is mainly within $10 \, {\rm ms}$ which satisfies the $100 \, {\rm Hz}$ running frequency condition of the NMPC.

During experiments, it is typically unfeasible to make the head tip of the cable $\bm{r}(0,t)$ and the quadrotor's CoM $\bm{p}_B$ exactly coincide. However, we assume the attitudinal motion of the quadrotor does not considerably affect the cable's motion. Based on this assumption, the translational accelerations of quadrotor CoM and cable head tip are considered approximately equal to each other $\Ddot{\bm{p}}_B \approx \bm{r}_{tt}(0,t)$, making~\eqref{eq:39} still valid.

\subsection{Cable model identification and validation}

\begin{figure}[t]
\centering
\includegraphics[width=7cm]{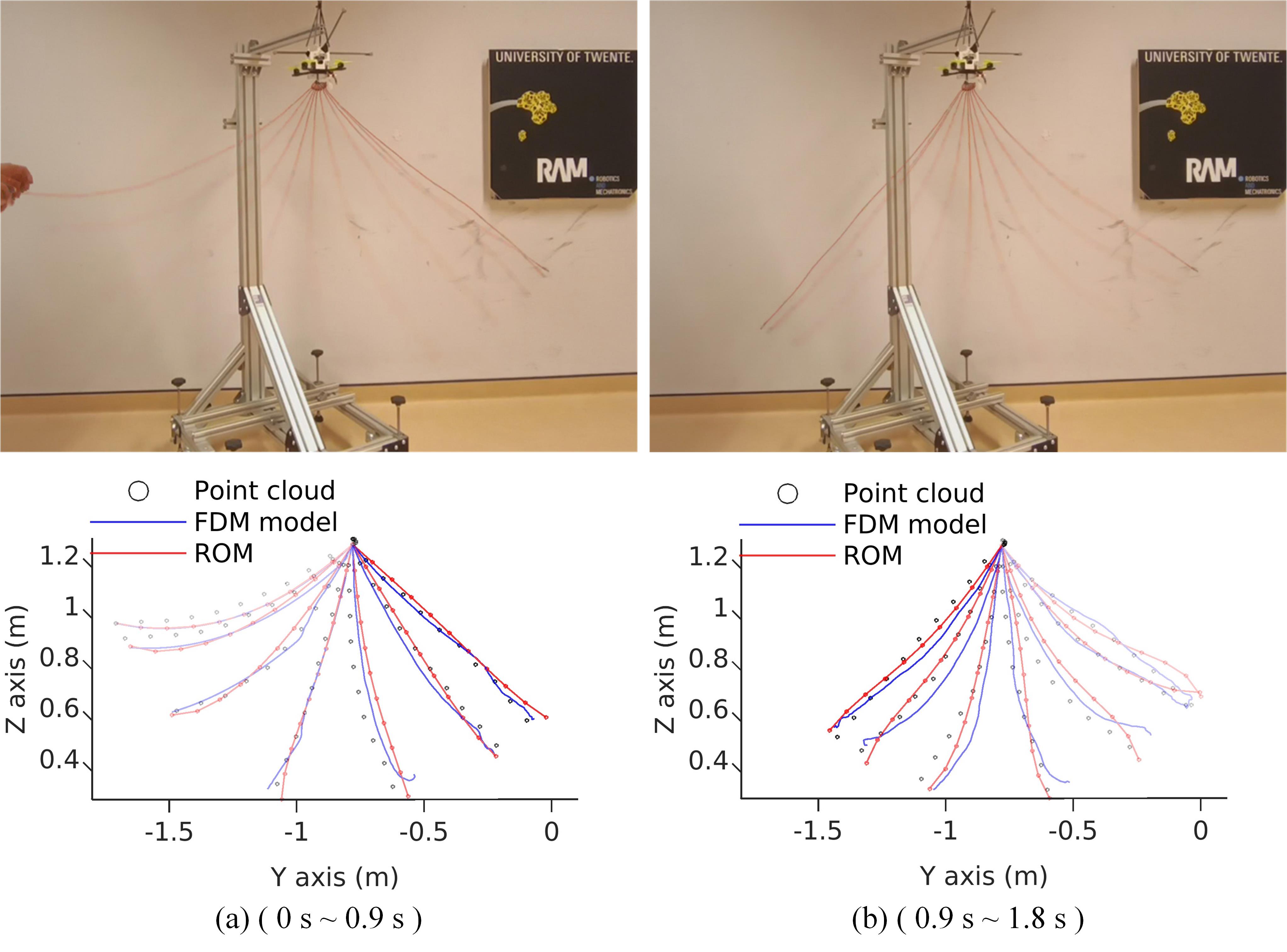}
\caption{\label{fig:19}\ys{Cable's motion sequence. Upper figures: a snapshot of the cable for 2 seconds after release. Lower figures: visualization of the evaluation of the real cable's point cloud (black dots), FDM model (blue line), and cable ROM during the identification experiment (red segmented line).}} 
\end{figure}

The parameters of the cable model~\eqref{eq:12} are required to formulate the NMPC. The cable's length $L= 1 \, {\rm m}$, cross-sectional area $A= 3.1416 \times 10^{-6} \, {\rm m^2}$, and density $\rho_c=3.0239 \times 10^{3} \, {\rm kg \cdot m^{-3}}$ have been directly measured. The values of coefficients including the aerodynamic drag coefficient $c_d$ and cable's Young’s modulus are obtained via the cable model identification experiment shown in Figure~\ref{fig:19}. 

During the identification experiment, the quadrotor is fixed, and the cable is stretched by lifting the tip. After that, the cable is released, and the point cloud of the cable is recorded during the free-swing motion. The recorded point cloud data is then used in an optimization problem to obtain parameter values:
\begin{equation}
\begin{array}{rrl}
     &\{ c_d, E \}={\rm arg} \, {\rm min}_{c_d, E} & 
     \sum\limits_{i=1}^{\frac{T_{\rm exp}}{t_{\rm exp}}} \sum\limits_{k=1}^{10}  
     || 
     \bm{\mathbbm{r}}_{\rm exp}^{k}
     ( i \cdot t_{\rm exp})  \\
     & & \quad \quad \quad -\bm{r}^{\frac{k h_r}{h_s}}(i \cdot t_{\rm exp}) || _2 \\
\end{array},
\label{re_eq:9}
\end{equation}
where $T_{\rm exp}= 7 \, {\rm s}$ is the cable point cloud data recording time duration in the experiment, $t_{\rm exp} = 0.01 \, {\rm s}$ is the data recording time step size. $\bm{\mathbbm{r}}_{\rm exp}^{k} \in \mathbbm{R}^3$ is the position of $(k+1)^{\rm th}$ tap along the cable (the order is from head tip to tail tip). The cable shape $\bm{r}$ is simulated via the FDM method (Sec. \ref{subs:fdm}) whose discrete interval number $N=50$, simulation step is $0.05 \, {\rm ms}$ and integral formulation is fourth-order Runge-Kutta method. The initial condition for the FDM is obtained as the equilibrium state of the cable when fixing the positions of two endpoints (head and tail tips). 

The solution of the optimal identification problem~\eqref{re_eq:9} is obtained as $c_d=0.0013648 \, {\rm N \cdot m \cdot s^2 \cdot kg^{-1}}$ and $E=1234623.7038 \, {\rm N \cdot m^{-2}}$. After the system identification via the FDM method, the PDEs~\eqref{eq:12} of the cable is simplified as ROM with truncated order $K=3$ via the POD method (Sec. \ref{sec:model_reduced}).Such a ROM is simulated with a fixed time step of $10 \, {\rm ms}$ by the fourth-order Runge-Kutta method.

\begin{figure}[t]
\centering
\includegraphics[width=8cm]{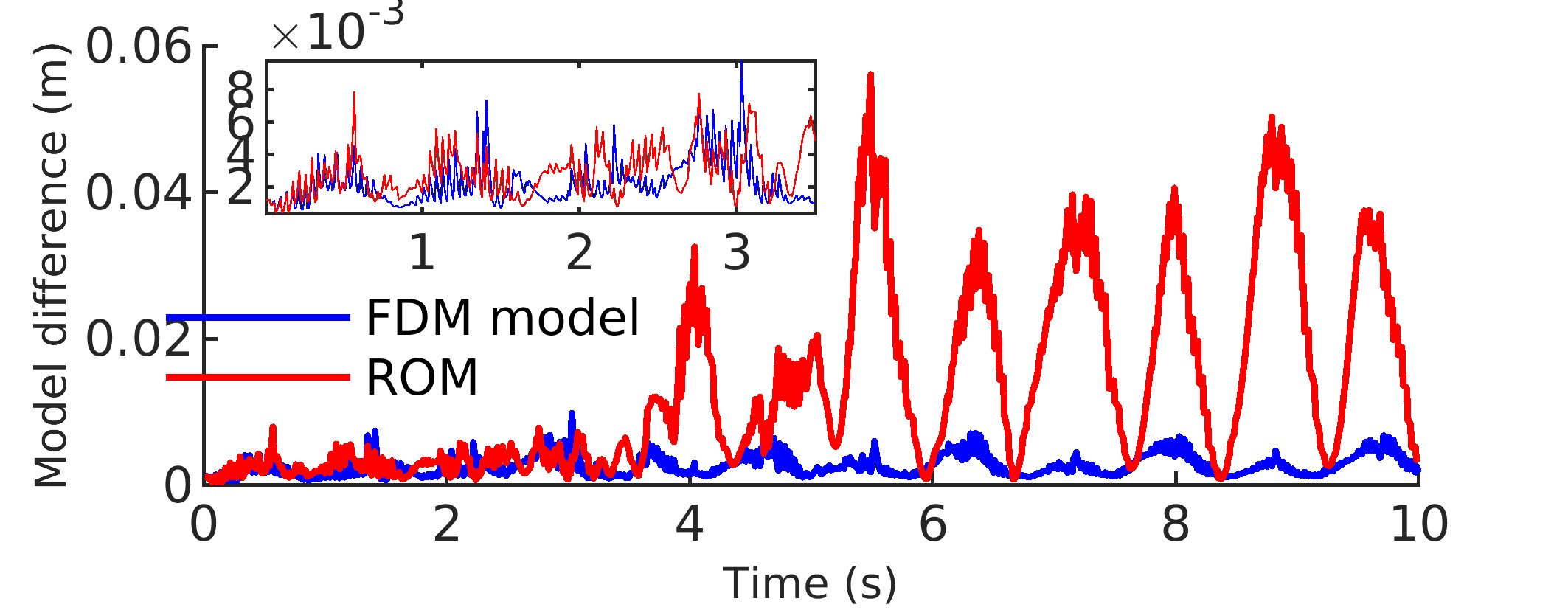}
\caption{\label{fig:20}\ys{Differences of FDM-based model and ROM, respectively, w.r.t. real cable}} 
\end{figure}

Figure~\ref{fig:19} and~\ref{fig:20} show the comparison between real cable (point cloud) and the two theoretical models (FDM model and ROM). Figure~\ref{fig:19} illustrates the motion sequence of the cable and also the corresponding cable motions obtained by the identified PDEs model (solved by FDM) and ROM. It was found that both the FDM model and ROM could describe the motion of the cable in the first $1.8 \, {\rm s}$ quite well.

For quantifying the differences between real cable motion and the motion obtained by the FDM model and ROM, respectively, two metrics are induced as in~\eqref{re_eq:13}:
\begin{equation}
\left\{
\begin{array}{c}
    E_{\rm exp}^{\rm FDM}=\frac{h_{\rm exp}}{L} \sum\limits_{k=1}^{\frac{L}{h_{\rm exp}}} {\norm{\bm{\mathbbm{r}}_{\rm exp}^{k}(t)-\bm{r}^{\frac{k h_{\rm exp}}{h_s}}(t)}}_2 \\
    E_{\rm exp}^{\rm ROM}=\frac{h_{\rm exp}}{L} \sum\limits_{k=1}^{\frac{L}{h_{\rm exp}}} {\norm{\bm{\mathbbm{r}}_{\rm exp}^{k}(t)- {\bm{r}}_{\rm ROM}(h_{\rm exp},t) }}_2  \\
\end{array}
\right\},
\label{re_eq:13}
\end{equation}
where  $\bm{r}^{\frac{k h_{\rm exp}}{h_s}}(t)$ obtained via FDM is the position of the cable's material point that refers to the corresponding reflective tape, and ${\bm{r}}_{\rm ROM}(h_{\rm exp},t)$ is the position of the same material point but obtained via ROM simulation. These two metrics could be interpreted as the average distance between the reflective tapes and their corresponding material points on the cable via the FDM model and ROM simulation, respectively.

From Figure~\ref{fig:20}, it can be seen that the values of the two metrics $E_{\rm exp}^{\rm FDM}$ and $E_{\rm exp}^{\rm ROM}$ are small (within $8 \, {\rm mm}$) for the first $3.5 \, {\rm s}$. After $3.5 \, {\rm s}$, the difference between real cable motion and the ROM rises dramatically (maximum value is around $6 \, {\rm cm}$). At the same time, the FDM model still describes the cable motion quite well. According to Figure~\ref{fig:20}, two conclusions could be derived: the proposed cable model described by PDEs in this work describes the real cable motion quite well in the long term; the cable's reduced order model described by ODEs mimics the real cable motion in the short term.

\subsection{Regulation problem}\label{real:stab}

The regulation problem of the real-world experiments is the same described in Sec.\ref{subs:stab}. In this subsection, the proposed NMPC and PID controllers are executed to stabilize the quadrotor-quadrotor system in two separate tests, where the cable is perturbed manually. 

The controller gain of the PID controller is set as follows: $\bm{{\rm K}}_p^r={\rm diag} \{ 15,15,30 \}$ and $\bm{{\rm K}}_d^r={\rm diag} \{ 10,10,20 \}$. And, the gains of the NMPC are set as follows: $\bm{{\rm Q}}={\rm diag} \{ 20,20,20,5,5,5,1,1,1,500,500,20,10,10,20,10,10,$ $10,0.1,0.1,0.1, 5,5,2\}$ and $\bm{{\rm R}}={\rm diag} \{ 0.05,0.05,0.05 \}$. The target shape of the cable is calculated as~\eqref{re:eq:pod:2}, where $\bm{r}(0,\cdot)=[0,0,1.5]^{\rm T}$.

\begin{figure*}[t]
\centering
\includegraphics[width=16cm]{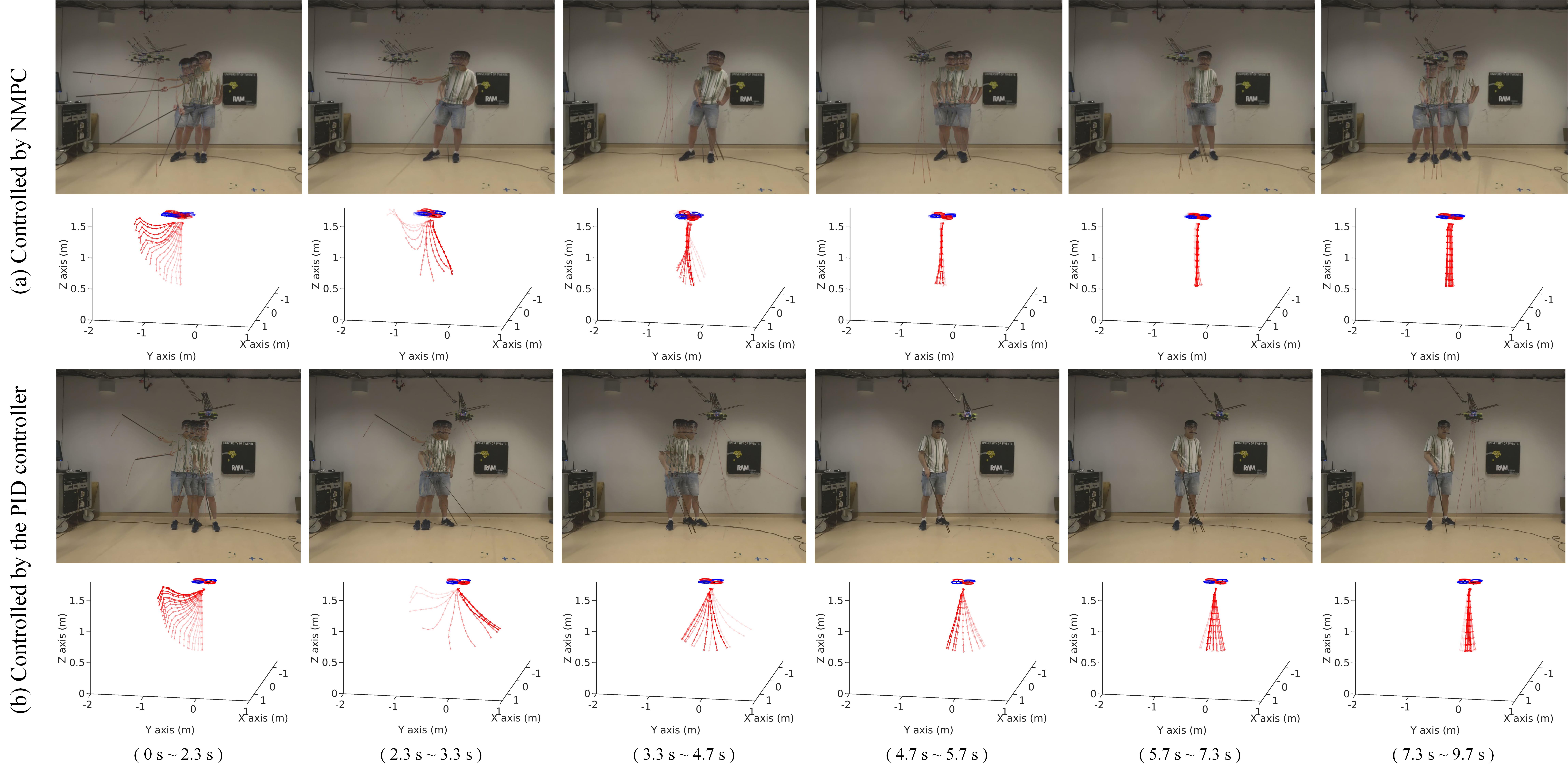}
\caption{\label{fig:21}\ys{Snapshot of the quadrotor-cable system controlled with the proposed NMPC (top) and PID (bottom) in the regulation scenario. The red curve is the measured cable shape and the red points on the curve are the positions of reflective tapes.}} 
\end{figure*}

Figures~\ref{fig:21}(a)-(b) show the motion sequence of the quadrotor-cable system controlled with the proposed NMPC and the PID controller, respectively. During these two tests, the cable is perturbed into a similar shape manually during the first $2.3 \, {\rm s}$. After $2.3 \, {\rm s}$, the cable is released, and the NMPC and PID controllers take the role of stabilizing the system. Comparing  Figure~\ref{fig:21}(a) and (b), it required around $3.5 \, {\rm s}$ for the proposed NMPC to stabilize the cable shape after releasing the cable, while for the PID controller, the stabilizing time is around $7.5 \, {\rm s}$, which means the NMPC superior performance than PID control for stabilizing the cable's shape. Additionally, from Figure~\ref{fig:21}(a) in the interval $(0 \, {\rm s} - 2.3 \, {\rm s})$, the quadrotor controlled by the proposed NMPC moves to follow the direction where the cable is stretched. This quadrotor's movement tries to reduce the perturbation on the cable's shape, and let the cable become more "straight" (close to the target shape). On the other hand, with the PID controller, the quadrotor remains in the original position.

\begin{figure*}
\centering
\includegraphics[width=16cm]{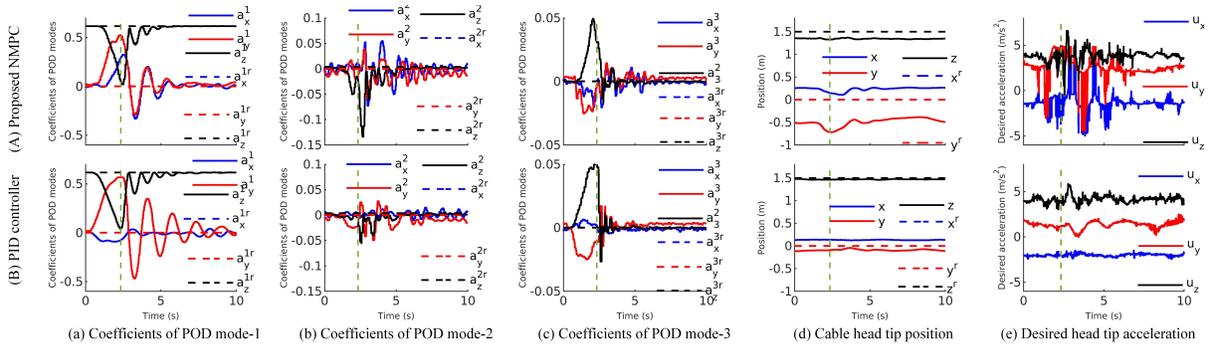}
\caption{\label{fig:22}\ys{Evolution of cable state in the regulation scenario using the proposed NMPC (top) and PID controller (bottom).}}
\end{figure*}

Figures~\ref{fig:22}(A)-(B) illustrate the evolution of the coefficients of the cable's POD mode with the two controllers and the outputs of the NMPC and the PID controller $\{ u_x, u_y, u_z \}$, which are the desired acceleration of the cable's head tip along three axes of the world frame (same as the quadrotor's translational acceleration for what explained earlier). Similar to the numerical experiment results in Sec.~\ref{subs:stab}, the NMPC controller successfully stabilizes the cable's first-order mode and the cable head tip position after $6 \, {\rm s}$. However, the PID controller fails to suppress the first-order mode vibrations even after double the time required for the NMPC. Additionally, different from numerical simulation, comparing \ref{fig:22}(A-b) and \ref{fig:22}(A-c) with \ref{fig:22}(A-b) and \ref{fig:22}(A-c), it is found that the residual vibration
amplitudes of the second-order and third-order modes are similar in the NMPC and PID controller cases; these behaviors may be caused by the noise  in actuation and sensor measurements, which are not considered in numerical experiments. 

The biggest difference between real-world and numerical experimental results in the regulation scenario is the cable head tip position. From Figure~\ref{fig:22}(A-d) and~\ref{fig:22}(B-d), it could be found that there are larger static position errors of the cable head tip with NMPC than with the PID controller. These static position errors are caused by external disturbance, while no disturbance is included in numerical cases. Ideally, according to~\eqref{re_eq:1}, the desired translational acceleration of the quadrotor $\Ddot{\bm{p}}_B^d$ (cable head tip acceleration ${\bm{r}_{tt}}^d(0,t)$) should be equal to $[0,0,0]^{\rm T}$ at the equilibrium state of the quadrotor when no model uncertainty and external disturbance exist. However, as shown in Figure~\ref{fig:22}(A-e) and~\ref{fig:22}(B-e), the desired accelerations of cable head tip (the same as translational acceleration of quadrotor) convergence to similar non-zero values in two cases, which means the quadrotor suffers similar external forces in these two tests. In this regulation experiment, the external forces are induced by the safety cable attached to the upper side of the quadrotor (shown in Figure~\ref{fig:21}), which pushes the quadrotor to the bottom-left direction.

Furthermore, according to Figure~\ref{fig:22}(A-d,e) and~\ref{fig:22}(B-d,e), it could be concluded that the quadrotor-cable system controlled by the proposed NMPC may have larger position errors on the cable head tip than controlled by the PID controller when the quadrotor suffered with the same external force. An intuitive explanation for this phenomenon is that the proposed NMPC adopts the non-disturbed cable model~\eqref{eq:47} to predict the behavior of the cable; however, to overcome the external disturbance, a non-zero controller output ${\bm{r}_{tt}}^d(0,t)$ is required, which will induce large cable shape error in the prediction horizon of NMPC. Hence, a trade-off exists between cable head tip position error and cable shape error when external distance is included. To solve this trade-off, a future solution could be adding an external disturbance term in the NMPC's surrogate model~\eqref{eq:37} and estimating or measuring this term online.

\subsection{Trajectory tracking}

The setup of this real-world experiment scenario is the same as the one in Sec.~\ref{subs:traj_track} but for the reference trajectory of the cable head tip $\bm{r}^{r}(0,t)$, which is here an eight shape expressed as~\eqref{re_eq:10}:
\begin{equation}
    \bm{r}^{r}(0,t)=[\sin{\frac{2\pi t}{10}}, \, \cos{\frac{2\pi t}{10}}, \, 1.5]^{\rm T}.
\label{re_eq:10}
\end{equation}

\begin{figure*}[t]
\centering
\includegraphics[width=13.5cm]{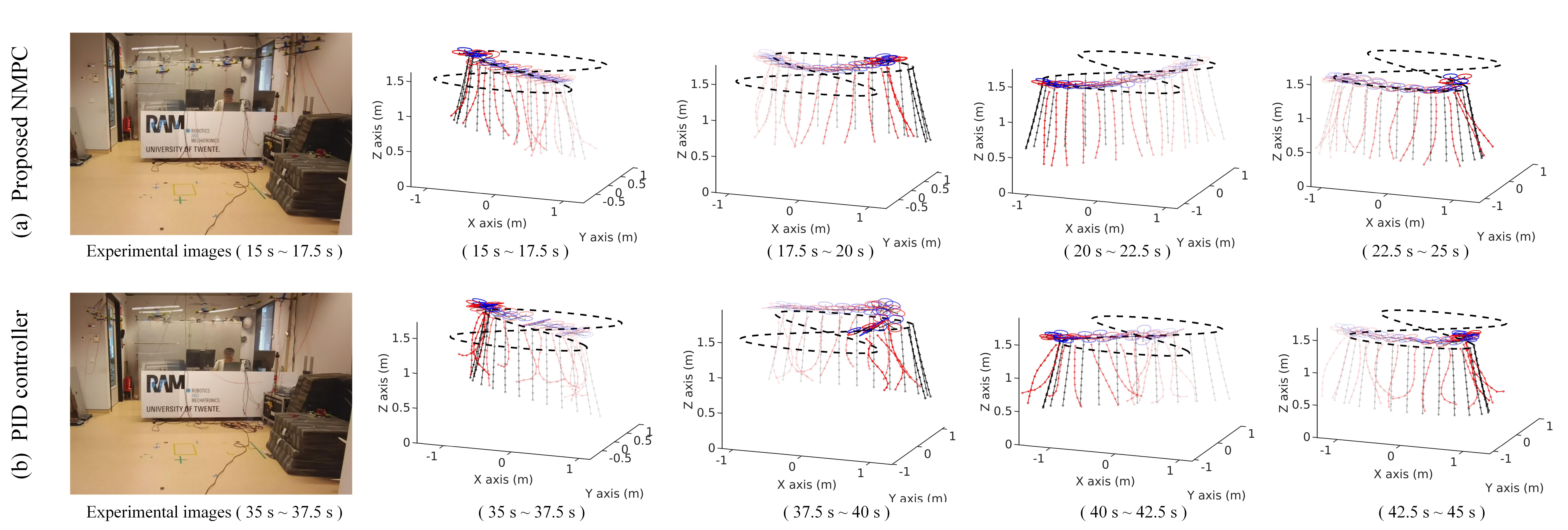}
\caption{\label{fig:24}\ys{Snapshot of the quadrotor-cable system controlled with the proposed NMPC (top) and PID controller (bottom) in the trajectory tracking scenario. The black dotted line refers to the reference of the cable head tip, the black solid curve is the reference cable shape and the red curve is the measured cable shape.}} 
\end{figure*}

The motion of the quadrotor-cable system in this scenario is illustrated in Figure~\ref{fig:24}. During this test, the quadrotor is controlled by the proposed NMPC to carry the cable following the reference shape at first $30 \, {\rm s}$, then the controller is switched to the PID controller. The controllers' gain used in the trajectory tracking scenario is the same as the last subsection Sec.~\ref{real:stab}.

By comparing Figure~\ref{fig:24}(a) and (b), one finds that with the proposed NMPC, the cable briefly follows its desired shape trajectory. On the contrary, when the quadrotor is controlled by the PID controller, the cable swings aggressively into different shapes.

\begin{figure}[t]
\centering
\includegraphics[width=8.5cm]{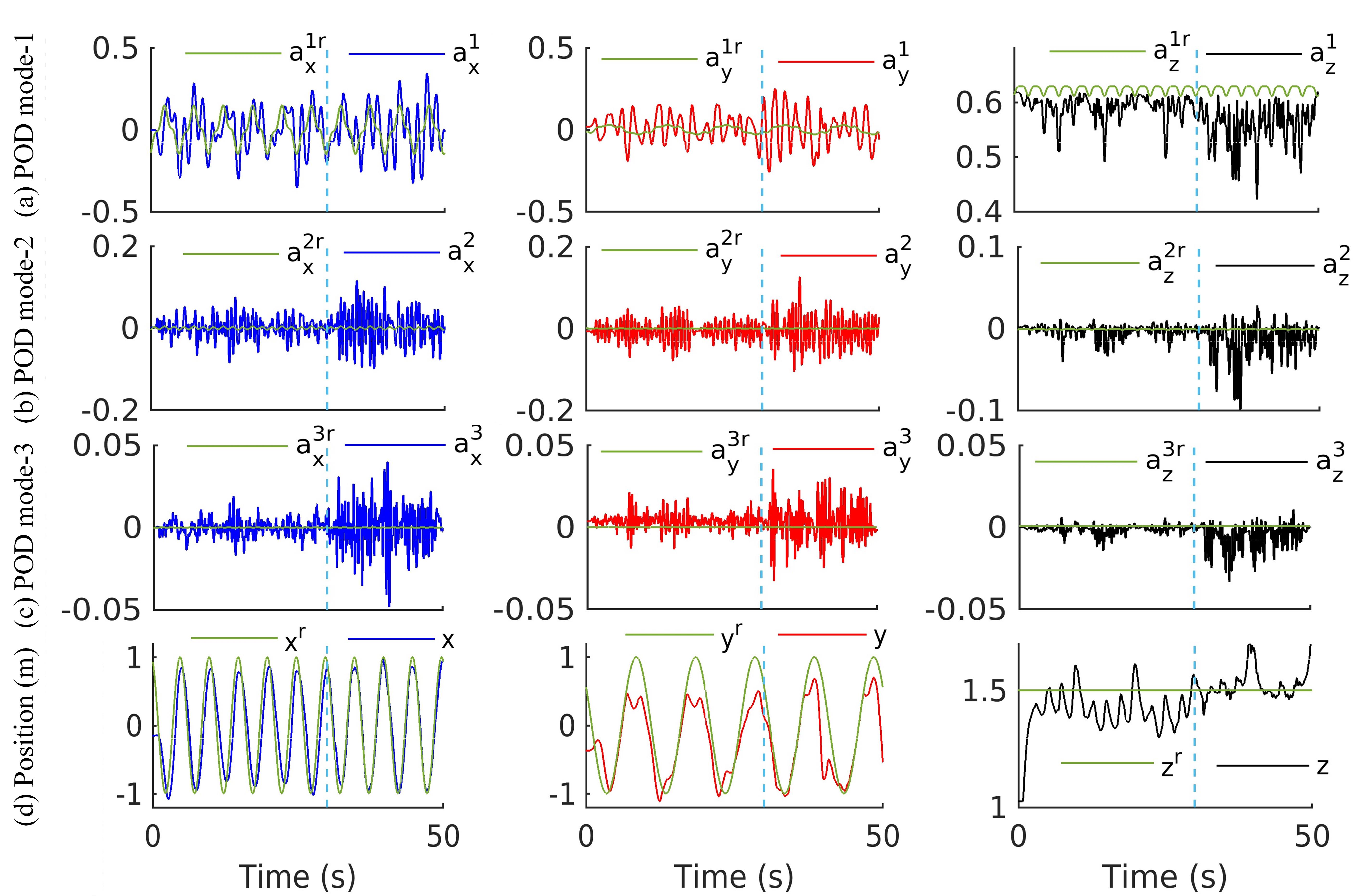}
\caption{\label{fig:25}\ys{Evolution of cable state during trajectory tracking. Blue dotted line indicates controller switch time: in the period $[0 \, {\rm s}-30 \, {\rm s}]$ the controller is the proposed NMPC. After $30 \, {\rm s}$, the controller is switched to the PID controller.}} 
\end{figure}

Figure~\ref{fig:25} illustrates the evolution of the coefficients of the cable’s POD mode and head tip position with the two controllers. For the first $30 \, {\rm s}$, the coefficients of the cable’s first three modes and the head tip position remain in the neighborhood of the desired reference. After that, when the controller is switched to the PID controller, the tracking performance becomes especially worse in the coefficients of the three modes than with the NMPC. However, due to the trade-off issue mentioned in the last subsection Sec.~\ref{real:stab}, the cable's head tip trajectory tracking by NMPC is slight worse than when the PID controller is used.

\subsection{Narrow window crossing}

\begin{figure}[t]
\centering
\includegraphics[width=8.5cm]{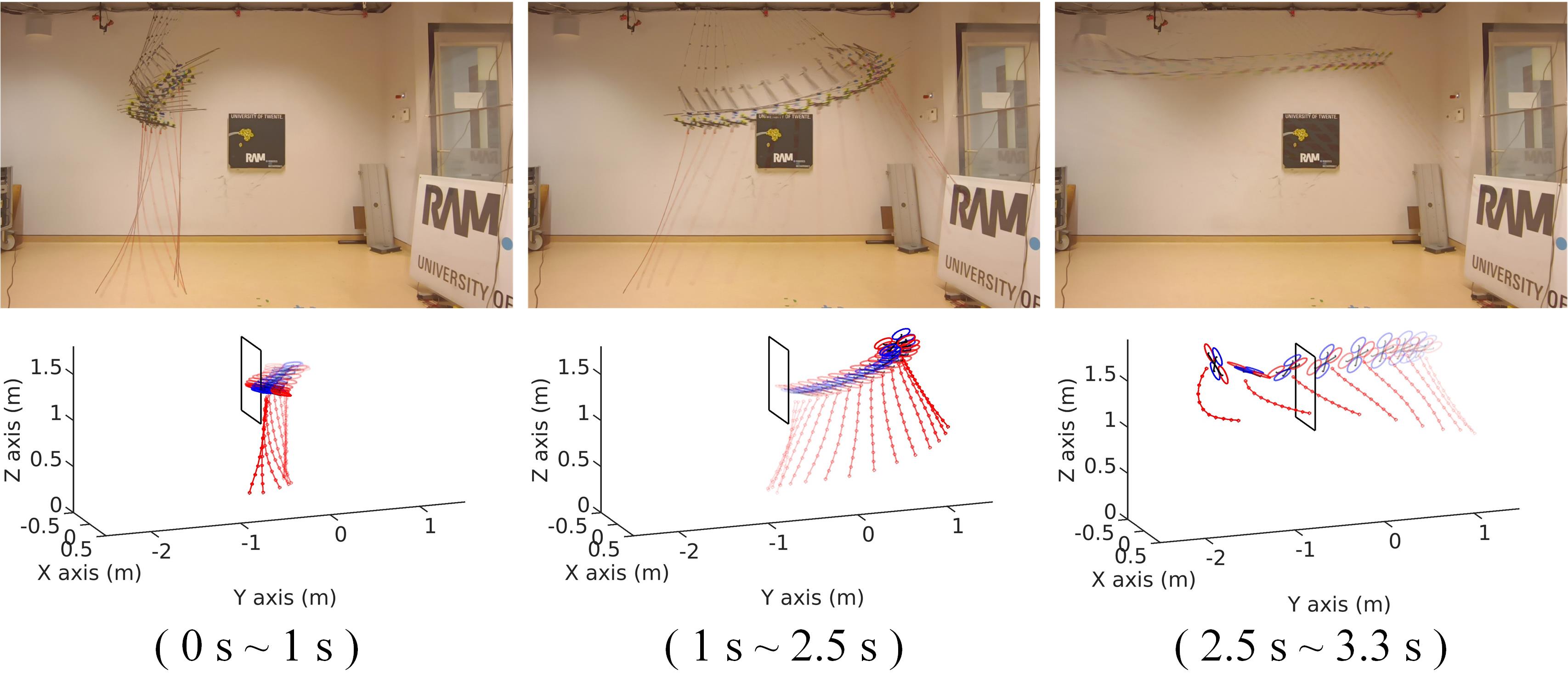}
\caption{\label{fig:26}\ys{Snapshots of the quadrotor-cable system during narrow window crossing controlled with the proposed NMPC. The vertical window is virtually visualized as a black rectangular.}} 
\end{figure}

Similarly as in Sec.~\ref{subs:window_crossing}, in this subsection, the quadrotor is required to carry the cable through a vertical window schematically represented in Figure~\ref{fig:26}. The collision constraint for the quadrotor and the cable is expressed as~\eqref{re_eq:11}, where $L_{q}=0.11 \, {\rm m}$ is the distance between the cable head tip attaching point and the quadrotor's CoM.
\begin{equation}
    \left\{
        \begin{array}{c}
            -0.3 < {r}^x(s,\cdot) < 0.3 \\
            (1+L_{q}) < {r}^z(s,\cdot) < (1.8-L_{q}) \\
        \end{array}
    \right\}
    \, \text{if}\, \, {r}^y(s,\cdot)=-0.7,
\label{re_eq:11}
\end{equation}

Instead of the one-dimensional motion planning used in the numerical experiment, the vertical motion of the cable head tip is also considered in the real-world experiment. In this case, the motion of the cable head tip is represented by polynomials:
\begin{equation}
    \bm{r}(0,t)=[a_0 + \sum\limits_{i=1}^{K_r} a_i t^i ,0,b_0 + \sum\limits_{i=1}^{K_r} b_i t^i]^{\rm T}
\label{re_eq:12}
\end{equation}

Then, the same procedures as Sec.~\ref{subs:window_crossing} are executed to obtain the reference trajectory of the quadrotor-cable system to cross the window without collision. The gains of the NMPC are set as follows: $\bm{{\rm Q}}={\rm diag} \{ 1,1,1,$ $1,1,1,1,1,1,10000,10000,1000,0.2,0.2,0.2,0.2,0.2,0.2,$ $0.1,0.1,0.1, 30,30,30\}$ and $\bm{{\rm R}}={\rm diag} \{ 0.1,0.1,0.1 \}$, which emphasizes the translational motion of the cable's head tip.

\begin{figure}[t]
\centering
\includegraphics[width=7cm]{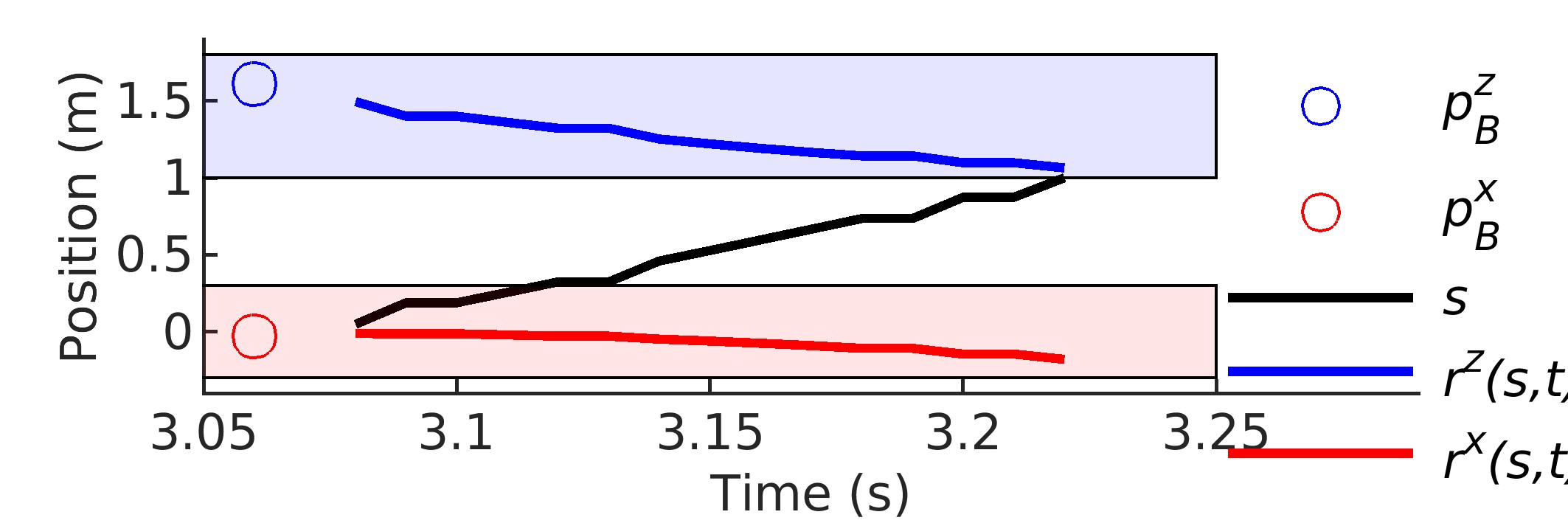}
\caption{\label{fig:28}Quadrotor and cable position within window plane}
\end{figure}

The motion sequence of the quadrotor-cable system for the window crossing task is illustrated in Figure~\ref{fig:26}. The crossing motion takes around $3 \, {\rm s}$: in the first second, the quadrotor flies in the bottom-left direction towards the window; then, from $1 \, {\rm s}$ to $2.5 \, {\rm s}$ the quadrotor moves in the opposite direction and swings up the cable; after $2.5 \, {\rm s}$ the quadrotor moves towards the left side and crosses the window fast (the maximum velocity and acceleration magnitudes are $6.3 \, {\rm m \cdot s^{-1}}$ and $10.8 \, {\rm m \cdot s^{-2}}$). Thanks to the cable's dynamic shape, even though the cable length is longer than the window's side, the cable and the quadrotor could cross the window without collisions. The quadrotor and the cable positions along $\bm{E_x}$ and $\bm{E_z}$ axes when they arrive at the window plane are shown in Figure.~\ref{fig:28}. It can be seen that both the quadrotor and the cable positions are within the window's area, which means the quadrotor-cable system successfully passes through the window without collisions.

\begin{figure}[t]
\centering
\includegraphics[width=8.5cm]{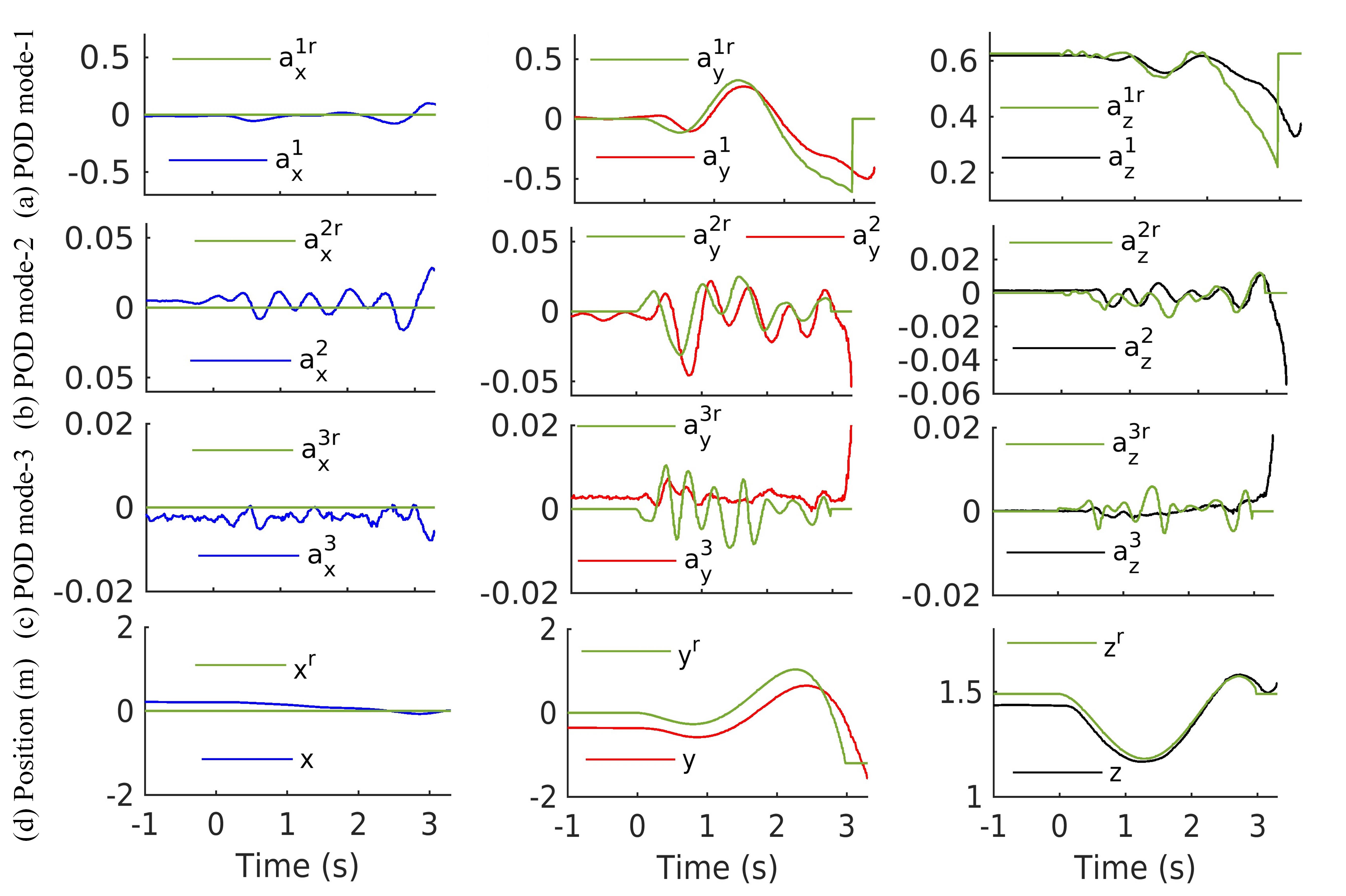}
\caption{\label{fig:27}\ys{Evolution of cable state in the window crossing}} 
\end{figure}

Figure~\ref{fig:27} shows the evolution of the coefficients of the cable POD modes and the position of the cable’s head tip during the crossing. It could be found that the coefficients of the first- and second-order POD modes of the cable and the position of the cable’s head tip follow their corresponding references. For the third-order POD mode, large tracking errors could be found in $1 \, {\rm s} - 2.5 \, {\rm s}$, which means the cable's ROM may not be precise enough to capture the third-order mode motion of the cable in this period.

\section{Conclusions}\label{sec:conclusions}

In this work, a novel PDE-based mathematical model for the system composed of a quadrotor carrying a flexible cable is presented. Then a Reduced-Order Model (ROM) model 
exploiting the proper orthogonal decomposition
is also proposed for control purposes. Numerical simulation results employing FDM-based discretization of the full PDE-based model show that the Reduced-Order Model (ROM) has the capability to predict the evolution of the system with sufficient accuracy and a much lower number of states. 
To perform cable shape trajectory tracking, we then proposed a novel NMPC scheme that uses the ROM in its prediction phase.

The proposed controller is numerically tested in actual control challenges simulating the system with the accurate model based on FDM in three different scenarios including regulation, cable shape trajectory tracking, and window crossing. 
The sensitivity of the control error to parametric variation is also numerically assessed. Real-world experiments are also executed to analyze the predictive capability of the ROM and test the performance of the ROM-based NMPC with a physical system in the same three scenarios.

In this work, the cable state estimation relies on a motion capture system. For an onboard implementation, onboard perception methods should be investigated.  




\bibliographystyle{plain}
\bibliography{ref}

\end{document}